\documentclass[10pt,twocolumn,letterpaper]{article}
\pdfoutput=1
\usepackage{cvpr}              

\usepackage[accsupp]{axessibility}  
\usepackage{graphicx}
\usepackage{amsmath}
\usepackage{amssymb}
\usepackage{booktabs}
\usepackage[pagebackref,breaklinks,colorlinks]{hyperref}
\usepackage{url}            
\usepackage{booktabs}       
\usepackage{multirow}
\usepackage{amsfonts}       
\usepackage{nicefrac}       
\usepackage{microtype}      
\usepackage{xcolor}         
\usepackage{graphicx}
\usepackage{bbold}
\usepackage{url}
\usepackage{pifont}
\usepackage{ulem}
\newcommand{\cmark}{\ding{51}}%
\newcommand{\xmark}{\ding{55}}%
\usepackage[capitalize]{cleveref}
\crefname{section}{Sec.}{Secs.}
\Crefname{section}{Section}{Sections}
\Crefname{table}{Table}{Tables}
\crefname{table}{Tab.}{Tabs.}

\def\cvprPaperID{7010} 
\def\confName{CVPR}
\def\confYear{2023}

\begin{document}

\title{Learning Situation Hyper-Graphs for Video Question Answering}

\author{Aisha Urooj Khan$^{1,3}$, Hilde Kuehne$^{2,4}$, Bo Wu$^2$, Kim Chheu$^5$,  \\Walid Bousselham$^4$,
Chuang Gan$^{2,6}$, Niels Lobo$^1$, Mubarak Shah$^1$\\
\\
$^1$ CRCV, University of Central Florida ,  $^2$ MIT-IBM Watson AI Lab, $^3$ Mayo Clinic, AZ \\
$^4$ Goethe University Frankfurt Germany,  
$^5$ Western Michigan University, $^6$ UMass Amherst\\
}
\maketitle

\begin{abstract}
   Answering questions about complex situations in videos requires not only capturing the presence of actors, objects, and their relations but also the evolution of these relationships over time. A situation hyper-graph is a representation that describes situations as scene sub-graphs for video frames and hyper-edges for connected sub-graphs and has been proposed to capture all such information in a compact structured form. 
In this work, we propose an architecture for Video Question Answering (VQA) that enables answering questions related to video content by predicting situation hyper-graphs, coined \textbf{S}ituation \textbf{H}yper-\textbf{G}raph based \textbf{V}ideo \textbf{Q}uestion \textbf{A}nswering (SHG-VQA). 
To this end, we train a situation hyper-graph decoder to implicitly identify graph representations with actions and object/human-object relationships from the input video clip. and to use cross-attention between the predicted situation hyper-graphs and the question embedding to predict the correct answer. 
The proposed method is trained in an end-to-end manner and optimized by a VQA loss with the cross-entropy function and a Hungarian matching loss for the situation graph prediction.
The effectiveness of the proposed architecture is extensively evaluated on two challenging benchmarks: AGQA and STAR. 
Our results show that learning the underlying situation hyper-graphs helps the system to significantly improve its performance 
for novel challenges of video question-answering tasks\footnote{Code will be available at \url{https://github.com/aurooj/SHG-VQA}}. 
\end{abstract}
\vspace{-10pt}
\section{Introduction}
\label{sec:intro}
Video question answering in real-world scenarios is a challenging task as it requires focusing on several factors including the perception of the current scene, language understanding, situated reasoning, and future prediction. 
 Visual perception in the reasoning task requires capturing various aspects of visual understanding, e.g., detecting a diverse set of entities~\cite{khan2021reason, khan2022_wsg_vlt}, recognizing their interactions, as well as understanding the changing dynamics between these entities over time. 
Similarly, linguistic understanding has its challenges as some question or answer concepts may not be present in the input text or video. 

\begin{figure}[t]
\begin{center}
  \includegraphics[width=\linewidth]{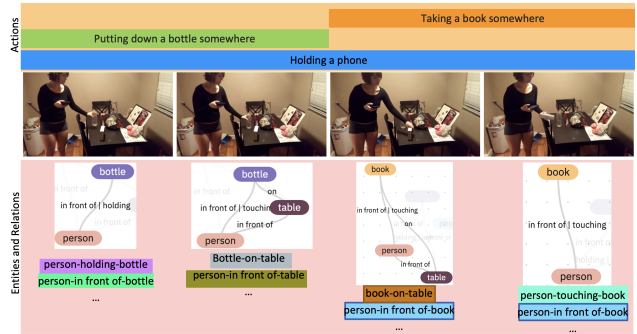}
 \end{center}
 \vspace{-7pt}
  \caption{The situation hyper-graph for a video is composed of situations with entities and their relationships (shown as subgraphs in the pink box). These situations may evolve over time. Temporal actions act as hyper-edges connecting these situations into one situation hyper-graph. Learning situation graphs, as well as temporal actions, is vital for reasoning-based video question answering.} \label{fig:teaser}
\end{figure}

Visual question answering, as well as its extension over time, video question answering, have both benefited from representing knowledge in graph structures, e.g., scene graphs~\cite{mao2018neuro,xnm}, spatio-temporal graphs\cite{damodaran2021understanding, yu2020ernie}, and knowledge graphs\cite{singh2019strings, marino2021krisp}.
Another approach in this direction is the re-introduction of the concept of ``\textit{situation cognition}" embodied in ``\textit{situation hyper-graphs}"~\cite{wu2021star}. 
This adds the computation of actions to the graphs that capture the interaction between entities. 
In this case, situations are represented by hyper-graphs that join atomic entities and relations (e.g., agents, objects, and relationships) with their actions (Fig. ~\ref{fig:teaser}).  
This is an ambitious task for existing systems as it is impractical to encapsulate all possible interactions in the real-world context. 

\begin{figure*}
\begin{center}
  \includegraphics[width=\linewidth]{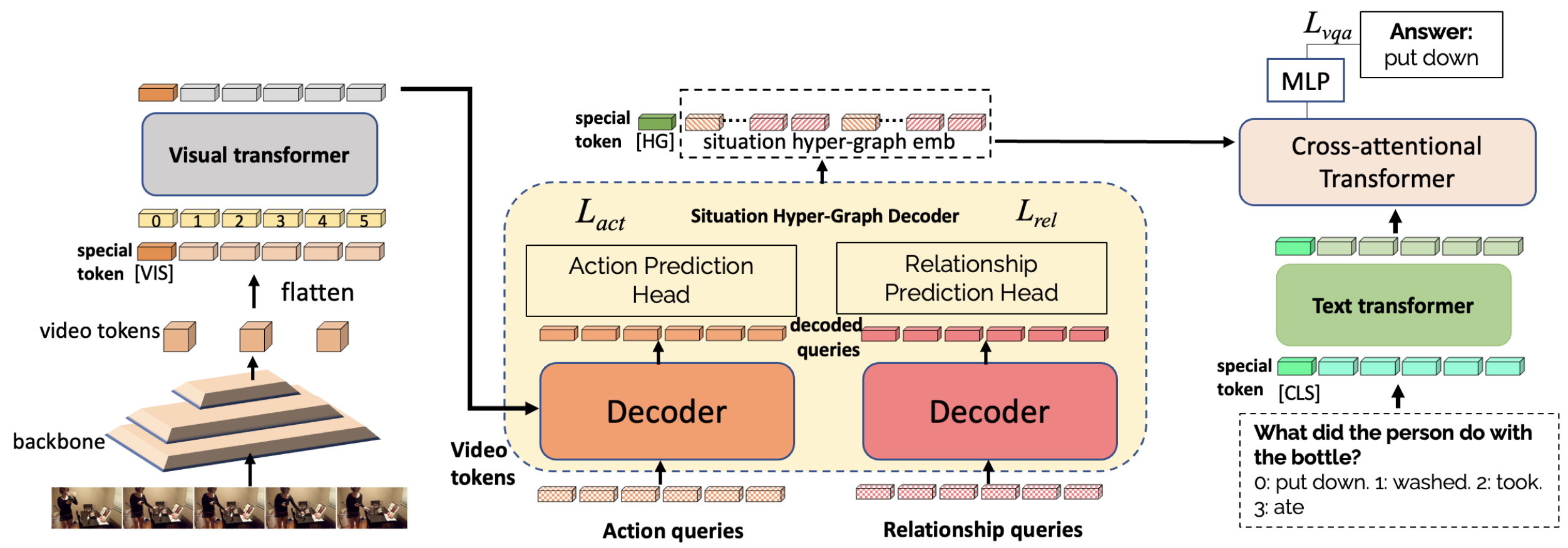}
 \end{center}
 \vspace{-10pt}
  \caption{The SHG-VQA architecture: we start with encoding the input video into spatio-temporal features using a pre-trained backbone. These video features are flattened into a sequence of tokens of length $T \times h \times w$ and position encoded to be further processed through a visual encoder. These encoded tokens are input to the action decoder to predict the set of atomic actions from the input action queries as well as to the relationship decoder which takes relationship queries as input along with the video tokens. The action decoder and relationship decoder output the situation graph embeddings. In the text branch, the question and the answer choices are composed into a sequence and passed through a text transformer to obtain encoded word embeddings; for open-ended VQA, only the question is passed to the text transformer. The generated hyper-graph along with the encoded text tokens are then used as input to a cross-attentional transformer and the combined representations are used to predict the correct answer with a classifier. Section~\ref{sec:loss-function} describes the losses and training objectives.} \label{fig:arch}
\end{figure*}

Recent work~\cite{kim2022pure} shows that transformers are capable of learning graphs without adapting graph-specific details in the architectures achieving competitive  or even better performance than sophisticated graph-specific models. Our work supports this idea by implicitly learning the underlying hyper-graphs of a video. Thus, it requires no graph computation for inference and
uses decoder’s output directly for cross attention module.
More precisely, we propose to learn situation hyper-graphs, namely framewise actor-object and object-object relations as well as their respective actions, from the input video directly without the need for explicit object detection or other required prior knowledge. 
While the actions capture events across transitions over multiple frames, such as \textit{Drinking from a bottle}, the relationship encoding actually considers all possible combinations of static, single frame actor-object, and object-object relationships as unique classes, e.g., in the form of \textit{person -- hold -- bottle} or \textit{bottle -- stands on -- table}, thus serving as an object and relation classifier. 
Leveraging this setup allows us to streamline the spatio-temporal graph learning as a set prediction task for predicting relationship predicates and actions in a Situation hyper-graph Decoder block. 
To train the Situation Graph Decoder, we use a bipartite matching loss between the predicted set and ground truth hyper-graph tokens. %
The output of the situation graph decoder is a set of action and relationship tokens, which are then combined with the embedding of the associated question to derive the final answer. 
An overview of the proposed architecture is given in Fig.~\ref{fig:arch}.
Note that, compared to other works targeting video scene graph generation, e.g., those listed in~\cite{Zhu2022SceneGraphGenerationSurvey}, we are less focused on learning the best possible scene graph, but rather on learning the representation of the scene which best supports the question answering task. 
Thus, while capturing the essence of a scene, as well as the transition from one scene to the other, we are not only optimizing the scene graph accuracy but also considering the VQA loss. 

We evaluate the proposed method on two challenging video question answering benchmarks: a) STAR~\cite{wu2021star}, featuring four different question types, interaction, sequence, prediction, and feasibility based on a subset of the real-world Charades dataset~\cite{Sigurdsson2016HollywoodIH}; and b) Action Genome QA (AGQA)~\cite{grunde2021agqa} dataset which tests vision focused reasoning skills based on novel compositions, novel reasoning steps, and indirect references. Compared to other VQA datasets, these datasets provide dense ground truth hyper-graph information for each video, which allows us to learn the respective embedding. Our results show that the proposed hyper-graph encoding significantly improves VQA performance as it has the ability to infer correct answers from spatio-temporal graphs from the input video. Our ablations further reveal that achieving high-quality graphs can be critical for VQA performance. 

Our contributions to this paper are as follows: 
\begin{itemize}
  \item We introduce a novel architecture that enables the computation of situation hyper-graphs from video data to solve the complex reasoning task of video question-answering;  
  \item We propose a situation hyper-graph decoder module to decode the atomic actions and object/actor-object relationships and model the hyper-graph learning as a transformer-based set prediction task and use a set prediction loss function to predict actions and relationships between entities in the input video; 
  \item We use the resulting high-level embedding information as sole visual information for the reasoning and show that this is sufficient for an effective VQA system. 
\end{itemize}

\vspace{-5pt}
\section{Related Work} \label{sec:related-work}
\paragraph{Video Question Answering:} 
Video question answering is an active area of research with efforts in insightful directions~\cite{zhong2022video,patel2021recent}, such as attention~\cite{jang2019video,li2019beyond,jiang2020divide}, cross-modal interactions~\cite{clipbert,seo-etal-2021-attend,yang2021just,yu2021learning,zellers2021merlot,zellers2022merlot, khan2020mmftbert}, hierarchical learning~\cite{peng2021progressive,dang2021hierarchical,hcrn}, and so on. A few other algorithms classes include modular networks~\cite{hcrn,dang2021hierarchical,xiao2021video}, symbolic reasoning~\cite{yi2019clevrer,nsvqa,wu2021star}, and memory networks~\cite{gao2018motion,fan2019heterogeneous}.
Several video QA benchmarks are introduced to evaluate this task from varying perspectives entailing description~\cite{maharaj2017dataset, DBLP:journals/corr/ZengCCLNS16,xu2017video, yu2019activitynet}, temporal reasoning~\cite{zhu2017uncovering, jang2017tgif,yi2019clevrer}, causal structures~\cite{yi2019clevrer,xiao2021next}, visual-language comprehension~\cite{tapaswi2016movieqa,lei2018tvqa,yang2021just}, relational reasoning~\cite{tapaswi2016movieqa}, and measuring social intelligence~\cite{zadeh2019social}. 
STAR~\cite{wu2021star} benchmark goes one step further providing a benchmark to perform diagnostic study at additional fronts such as predicting future interactions and feasibility of next possible actions in the unseen future. Existing approaches on the benchmark either rely on object features~\cite{ren2015faster} as nodes explicitly modeling their interactions through a message passing mechanism~\cite{lcgn,wu2021star}, or benefit from efficient input sampling strategies during training to learn robust visual representations~\cite{clipbert,hcrn}. We, however, take a different approach and focus on inferring the underlying semantic graph structure in the video and use it for QA reasoning. Our method is independent of using pretrained object detectors and uses a simple approach to learn to infer the sets of relationship predicates as well as actions for each frame. The model is trained end-to-end with the frozen backbone.
\vspace{-5pt}
\paragraph{Graph-based VQA:}
Another related line of work is graph-based VQA methods~\cite{li2019relation, kant2020spatially}. Some of them work on object features extracted from a pretrained detector (e.g., Faster-RCNN~\cite{ren2015faster})~\cite{lcgn,liu2021hair,seo2021look,dang2021hierarchical,xiao2021video}, while some operate on frame-level~\cite{clipbert,yu2021learning,jiang2020reasoning} (e.g., ResNet), clip-level features~\cite{peng2021progressive,seo-etal-2021-attend,yang2021just}(e.g., C3D, S3D) or transformer-based backbones~\cite{liu2021video,dosovitskiy2020image,qian2022scene}. We focus on predicting atomic actions, and relationship triplets from the frame-level or clip-level features directly instead of an explicit graph. 
\paragraph{Scene Graph Generation:} The proposed formulation of situation hyper-graph deviates from existing scene-graph generation approaches~\cite{Zhu2022SceneGraphGenerationSurvey, wu2021star, chang2021comprehensive} as we do not require object detections as input or object level supervision, nor do we model it in an explicit graph structure. Our goal is simple: given an input video, predict all object-relation-object and actor-relation-object triplets as well as associated actions for each video frame. Our intuition is that forcing the model to predict these predicates will drive the system to learn a latent graph structure for visual input. The predicted situation hyper-graph is treated as an abstract video representation and used for VQA reasoning in the next step.

\section{The SHG-VQA Model} \label{sec:method}
Situation video question answering has three essential steps 1) the visual recognition capacities of visual entities, their relationships, actions, and how these transition over time, 2) the language understanding capacity to the questions, 3) the question-guided reasoning process over the representation learned in the first step. Sub-optimal performance at any of these steps will affect the overall task performance.
A key problem here is that capturing the visual structures directly from raw data, e.g., in the form of features often results in a rather noisy signal and does not provide suitable input for high-level language-guided reasoning. 
To overcome this mismatch, we propose to learn the implicit structure of the visual input as an intermediate step between learning the video representation and question-based reasoning. Forcing the model to learn to predict this implicit structure (actions, relations between entities) not only improves the video representation but also acts as a lightweight, high-level representation of the video content and can be used for the VQA task. We illustrate our architecture in Fig. \ref{fig:arch}.

\subsection{Input Processing} \label{sec:problem-definition}
Given the input video and the question, we encode the inputs as described below:


\noindent \textbf{Question Encoder:} \label{sec:question-enc}
The question is first tokenized into word tokens using a wordPiece tokenizer. These word tokens along with the special class token $[CLS]$ are the inputs of an embedding layer. The output word embeddings from this layer are input to a transformer encoder that encodes each word using multi-head self-attention between different words at each encoder layer. 
 



\noindent \textbf{Video Encoder:} \label{sec:video-enc}
Let $V \in \mathbb{R}^{T \times H \times W \times 3}$ be the input video clip, where $T$ is the clip length with $3$ color channels, height $H$ and width $W$. First, we extract video features $x_{V}\in \mathbb{R}^{T \times h \times w \times d_x}$ using a convolutional backbone with $d_x$ being the feature dimension, $h$ and $w$ are the reduced feature's height and width. As transformers process sequential input data, the video features $x_{V}$ are flattened into a sequence (of size $\mathbb{R}^{Thw \times d_x}$) and reduced to dimension $d$ through a linear layer. Then, we append a trainable vector of dimension $d$ for a special class token $[VIS]$ to this sequence of video features at index 0. These features are combined with position encodings  and input to a transformer-based video encoder $V_e$. 

The output features of the video are forwarded to the action decoder and relationship decoder to infer the situation graph capturing the action information, as well as the entities and their relationships. 
The output of the last layer of both decoders is then combined and augmented with frame positions, forming the final hyper-graph embedding. We further attach a randomly initialized class token $[HG]$.
These situation graph embeddings are then input to a multi-layered cross-attentional transformer encoder for more fine-grained interaction between the question words and semantic knowledge extracted from the video in the form of a graph. The output features corresponding to $[CLS]$ and $[HG]$ tokens are input to an answer classifier to produce an answer. 


\subsection{Situation hyper-graph Generation} \label{sec:graph-generation}
Real-world video understanding relies on the scene understanding including changing relationships between the objects over time and the evolving actions. Therefore, we represent the given video as a ``situation graph" denoted by $G$ which describes actions and relationships between objects. Let $G = (V, E)$, where $V$ is the set of vertices representing all possible entities in the dataset, and $E$ is the set of edges representing all possible relationships between each pair of entities. Given an input video, we want to learn a situation graph $g_t= (v_t, e_t), v_t \in V, e_t \in E$ for each time step $t \in \{1, 2, ..., T\}$ in the video, that captures the entities (objects, actors), their relationships as well as the associated actions that are present in that frame. The hyper-graph for one video is thus represented by the set of situation graphs $\mathcal{G}=\{g_1, ... g_T\}$. An action may comprise multiple relationships and objects. For each frame, we further have a set of actions $A_t = {a_1, a_2, ..., a_N }$. The set of actions for the full video is then given as $\mathcal{A}=\{A_1, ...A_T\}$. Note that the actions are predicted in addition to the graph structure and that both will be merged in a \textit{situation hyper-graph} embedding in a separate step after the decoder block.  
Rather than predicting the set of vertices and edges in the graph $g_t$, we propose to predict the graph structure by formulating this graph prediction as follows:

Let $x_{V_e} \in \mathbb{R}^{T' \times h \times w \times d}$ be the encoded video features, where $T'$ denotes the temporal length of encoded features, $h$ and $w$ are spatial dimensions, and $d$ is the feature's dimension. A relationship predicate $p$ describes interactions between entities by triplet tokens $object-relation-object$ or $actor-relation-object$.
We propose to predict the set of relationship predicates $\textit{R}$ and the set of atomic actions $\textit{A}$ occurring in each video frame. This is intuitive because the ability to predict the atomic actions and relations between entities for each time step benefits the high level reasoning tasks such as video question answering in this work. 
Therefore, for each time step $t \in \{1, 2, ..., T\}$ in the video, we predict the set of relationship predicates denoted by  $R_t$ in each video frame where $R_t = \{p_1, p_2, ..., p_M\}$ and $p_i$ is the $i^{th}$ predicate between two entities, representing the vertices $v_x$ and $v_y$ (object-object or actor-object) and the relation resp. edge $e_i$ represented as $<v_x, e_i, v_y>$, $M=|R_t|$ is the relationship set size. Additionally, we predict the set of $N$ actions $A_t$ for each time step $t$ where $A_t = \{a_1, a_2, ..., a_N\}$ and $a_j$ is the $j^{th}$ action occurring at time step $t$. $N=|A_t|$ is the actions set size for each step $t$. 
$M$ and $N$ are hyperparameters in our system. 




To obtain actions and relation predicates from the video, we use a transformer decoder that takes video features as memory to learn action queries and relation queries. 
\begin{figure}[t]
\begin{center}
  \includegraphics[width=\linewidth]{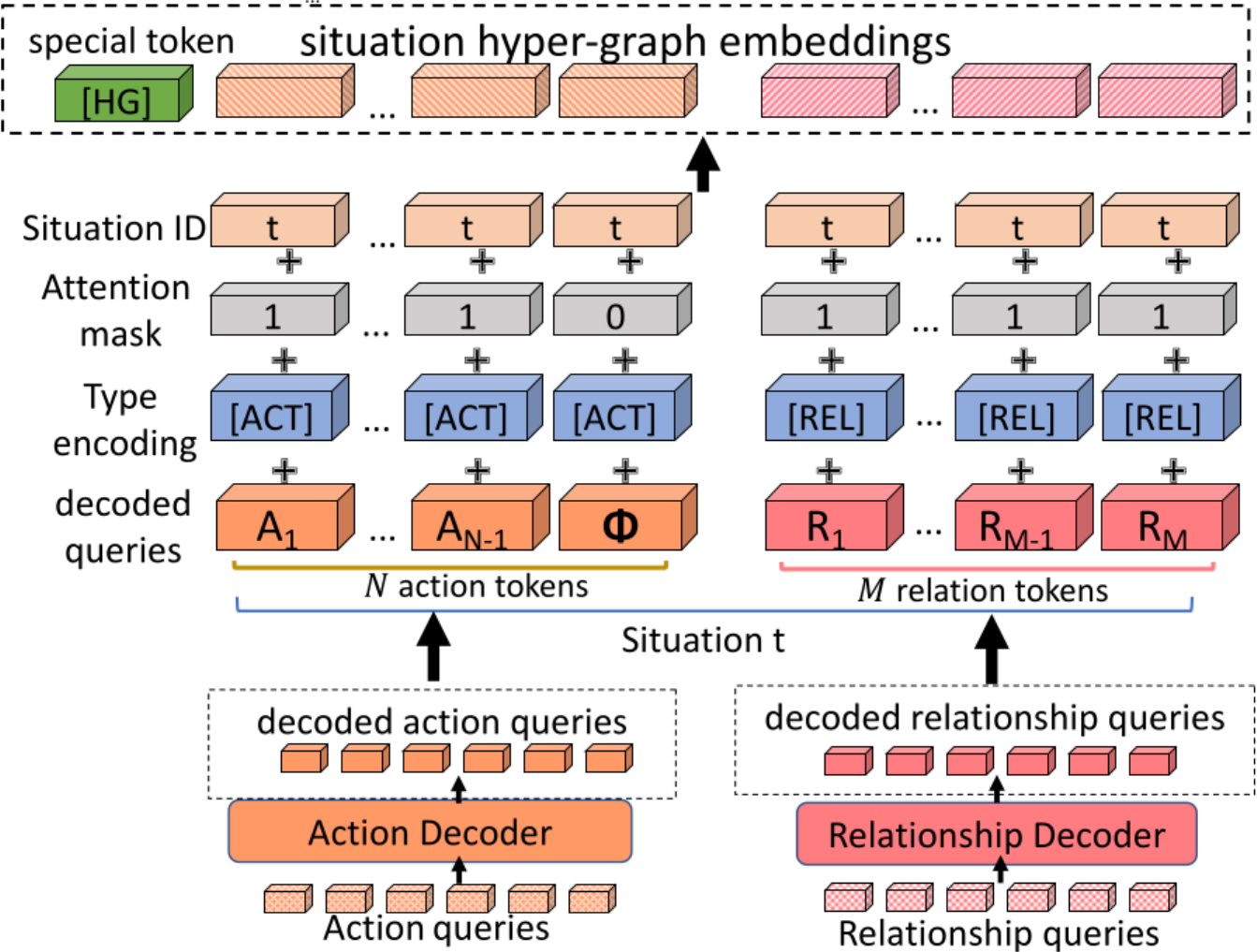}
 \end{center}
 \vspace{-10pt}
  \caption{Situation hyper-graph embeddings: 
  We start with decoded queries from action and relationship decoder. Then we add type encoding vectors $[ACT]$ and $[REL]$ for actions and relationships, attention masks, and an embedding vector for the situation ID ($t \in \{1,...,T\}$). 
 The sums are input to the cross-attentional module. See section \ref{sec:hypergraph} for details).} \label{fig:graph-embed}
  \vspace{-10pt}
\end{figure}
\vspace{-10pt}
\subsubsection{Prediction Head} \label{sec:prediction-heads}
The decoded output embeddings for action and relationship queries are input to respective prediction heads.
 Prediction heads use a 2-layer feed forward network (FFN) with GELU activation and LayerNorm.

Considering that actions and relationships at a given time step are permutation-invariant, we use optimal bipartite matching between predicted and ground truth actions/relationships. An optimal bipartite matching between the predicted classes and ground-truth labels is the one with the minimum matching cost. Once the optimal matching pairs have been obtained, we use a Hungarian loss function to optimize for ground truth classes~\cite{detr}. See section \ref{sec:loss-function} for details of the loss function.

\begin{table*}
  \caption{Results on AGQA dataset for different question types. The best results are shown in \textbf{bold} font. Numbers are reported in percentages.
  }
  \label{tab:agqamain-table}
  \renewcommand{\arraystretch}{.9}
  \centering \scriptsize \setlength{\tabcolsep}{.25\tabcolsep}
  \begin{tabular}{lcccccccccccccccccccccc}
    \toprule
& \multicolumn{8}{c}{Reasoning} & & \multicolumn{3}{c}{Semantic} && \multicolumn{5}{c}{Structure} && \multicolumn{3}{c}{Overall} \\
\cmidrule{2-9} \cmidrule{11-13} \cmidrule{15-19}  \cmidrule{21-23}
Method & obj-rel & rel-action & obj-action & superlative & sequencing & exists & duration & activity && obj & rel & action && query & compare & choose & logic & verify && binary & open & all \\
\midrule
PSAC~\cite{PSAC}  & 37.84 & 49.95 & 50.00 & 33.20 & 49.78 & 49.94 & 45.21 & 4.14 && 37.97 & 49.95 & 46.85 && 31.63 & 49.49 & 46.56 & 49.96 & 49.90 && 48.87 & 31.63 & 40.18 \\
HME~\cite{HME} & 37.42 & 49.90 & 49.97 & 33.21 & 49.77 & 49.96 & 47.03 & 5.43 && 37.55 & 49.99 & 47.58 && 31.01 & 49.71 & 46.42 & 49.87 & 49.96 && 48.91 & 31.01 & 39.89\\
HCRN~\cite{hcrn} & 40.33 & 49.86 & 49.85 & 33.55 & 49.70 & 50.01 & 43.84 & 5.52 && 40.33 & 49.96 & 46.41 && 36.34 & 49.22 & 43.42 & 50.02 & 50.01 && 47.97 & 36.34 & 42.11 \\
\midrule
SHG-VQA & \textbf{46.42} & \textbf{60.67} & \textbf{64.63} &\textbf{38.83} & \textbf{62.17} & \textbf{56.06} & \textbf{48.15} & \textbf{10.12} && \textbf{47.61} & \textbf{56.19} & \textbf{53.83} && \textbf{43.42} & \textbf{60.68} & \textbf{47.76} & \textbf{52.86} & \textbf{56.63} && \textbf{55.04} & \textbf{43.42} & \textbf{49.20} \\

\bottomrule
\end{tabular}
\end{table*}
\subsubsection{Situation hyper-graph Embedding} \label{sec:hypergraph} 
For the decoded queries of actions and relationships (referred as graph token embeddings), \textit{situation hyper-graph} embeddings are constructed in order to be used with question features for video question answering. 
First, these action and relationship graph embeddings are combined for each time step \textit{t} representing a situation at $t$.
Then, we add token type embedding $[ACT]$ for actions and $[REL]$ for relations to their respective graph embeddings. A situation ID (or frame  position $t$) embedding is also added to these embeddings. An additional attention mask is used to differentiate actual tokens and padded tokens (no-class token $\phi$) at training time. At inference time, no attention mask is used as we do not employ any information about the graph at test time. Finally, we add a special class token $[HG]$ to this sequence of features. See Fig.~\ref{fig:graph-embed} for visualization.

\subsection{Cross-attentional Transformer Module} \label{cross-attention-module}
The situation hyper-graph embeddings obtained at previous step (section~\ref{sec:hypergraph}) are input along with the question to a cross-attentional transformer module which allows fine-grained computation between the question features and the graph features. A standard co-attentional transformer module is used for cross-attention between the two sequential feature inputs.
The feature outputs corresponding to the $[HG]$ token and $[CLS]$ token from the cross-attentional transformer block are fed to a feed-forward network (FFN) for answer prediction. 



\subsection{Learning Objective} \label{sec:loss-function}
The SHG-VQA model is trained with the following training objective i.e., 
\begin{equation} \label{eq:2}
    L = L_{act} + L_{rel} + L_{vqa} 
\end{equation}
where $L_{act}$ and $L_{rel}$ are the set prediction loss terms for predicting the action set and relationship set for the video, and $L_{vqa}$ is the cross-entropy loss over the predicted situation graph and question.
\vspace{-5pt}
\paragraph{Actions and relationships set prediction loss:}
The situation graph prediction module infers fixed sets of sizes $|N| \times T$ actions and $|M| \times T$ relationships for $T$-length video clip in a single pass through the action decoder and relationship decoder respectively. 
We modify the set prediction loss used in \cite{detr} as follows.
Let $A$ be the set of ground-truth actions and $\hat{A}=\{\hat{a_i}\}^{|N|\times T}_{i=1}$ be the predicted set of actions. 
In a scenario where $\hat{A}$ is larger than the set of actions present in the video, a special class $\phi$ (no class) is padded to the ground truth set $A$. We obtain a bipartite matching between ground-truth and predicted set for each timestep $t$ as follows:
\begin{equation} \label{eq:3}
    \hat{\sigma_{a}} = \sum_{t}^{T} argmin_{\sigma_{t} \in \zeta_{|N|}} \sum_{i}^{|N|} \mathcal{L}_{match} (a_{t_{i}}, \hat{a}_{\sigma_{t}(i)})
\end{equation}

\begin{table*}
  \caption{Results on AGQA's novel compositions test metric.}
  \label{tab:agqa-novel-comp}
  \renewcommand{\arraystretch}{.9}
  \centering \scriptsize \setlength{\tabcolsep}{.7\tabcolsep}
  \begin{tabular}{lccccccccccccccccccc}
  \toprule
  & \multicolumn{3}{c}{Sequencing} && \multicolumn{3}{c}{Superlative} && \multicolumn{3}{c}{Duration} && \multicolumn{3}{c}{Obj-relation} && \multicolumn{3}{c}{Overall}\\
  \cmidrule{2-4} \cmidrule{6-8} \cmidrule{10-12} \cmidrule{14-16} \cmidrule{18-20}
  Method & B & O & All && B & O & All && B & O & All && B & O & All && B & O  & All \\
  \midrule
  PSAC~\cite{PSAC} & 49.19 & 29.33 &40.96 && 45.23 & 17.76 & 33.32 && 47.89 & 34.84 & 42.06 && \textbf{43.76} & 0.01 & 24.28 && 46.49 & 19.34 & 34.71\\
  HME~\cite{HME} & 49.33 & 28.06 & 40.53 && 44.06 & 13.8 & 30.95 && 48.45 & 34.72 & 42.31 && 39.58 & 0.00 & 21.96 && 45.42 & 17.17 & 33.15\\
  HCRN~\cite{hcrn} & 48.31 & 30.00 & 40.73 && 45.12 & 17.30 & 33.06 && 46.15 & 39.11 & 43.01 && 37.15 & 2.86 & 21.88 && 44.88 & 20.12 & 34.13\\
  \midrule
  SHG-VQA & \textbf{50.88} & \textbf{38.59} & \textbf{45.79} && \textbf{51.14} & \textbf{23.64} & \textbf{39.25} && \textbf{51.84} & \textbf{49.21} & \textbf{50.66} && 39.73 & \textbf{6.23} & \textbf{24.82} && \textbf{49.07} & \textbf{26.68} & \textbf{39.37}\\

\bottomrule
\end{tabular}
\end{table*}

Where, $\sigma_{t}$ is a permutation of N elements for frame t, $\mathcal{L}_{match} (a_{t{i}}, \hat{a}_{\sigma_{t}(i)})$  is a pair-wise matching cost between $i^{th}$ ground-truth action label in $t^{th}$ frame i.e., $a_{t_{i}}$ and a predicted action label at index $\sigma_{t}(i)$.
This optimal assignment for each step $t$ is computed using the Hungarian algorithm and the cost is summed over all $T$ steps. The proposed set prediction loss takes into account only the class predictions for all video frames with no bounding box ground truths being used, different from the original set prediction loss used in \cite{detr} for object detection in images. Let $\hat{p}(c_{t(i)})$ be the class probability for the action prediction at $\sigma_{t}(i)$, $\mathcal{L}_{match} (a_{t_{i}}, \hat{a}_{\sigma_{t}(i)})$ would be $-\mathbb{1}_{\{c_{t(i)} \neq \phi \} }\hat{p}_{\sigma_{t}(i)}(c_{t(i)})$. 
After we obtain a one-to-one optimal matching between ground-truth and predicted set items without duplicates at each time step, we can compute the loss between the matched pairs using a Hungarian loss as follows:
\begin{equation} \label{eq:4}
   \mathcal{L}_{act}(a, \hat{a}) = \sum_{i=1}^{|N|\times T} -\log \hat{p}_{\hat{\sigma}(i)} (c_{i}) 
\end{equation}

Likewise, $R$ is the set of ground-truth relations and $\hat{R}=\{\hat{p_i}\}^{|M|\times T}_{i=1}$ denotes the predicted relationships. $L_{rel}$ is formulated as follows:

\begin{equation}
    \hat{\sigma_{p}} = \sum_{t}^{T} argmin_{\sigma_{t} \in \zeta_{|M|}} \sum_{i}^{|M|} \mathcal{L}_{match} (p_{t_{i}}, \hat{p}_{\sigma_{t}(i)})
\end{equation}

\begin{equation} \label{eq:5}
   \mathcal{L}_{rel}(p, \hat{p}) = \sum_{i=1}^{|M|\times T} -\log \hat{p}_{\hat{\sigma}(i)} (c_{i}) 
\end{equation}

where $\hat{\sigma}(i)$ is the optimal matching obtained in the previous step for each frame. The inferred  graph is input to the cross-attentional transformer module along with the question and the answer choices as explained in section \ref{sec:hypergraph}. The proposed network is trained in an end-to-end manner.



\begin{table}[h]
 \renewcommand{\arraystretch}{.9}
  \centering \scriptsize \setlength{\tabcolsep}{.6\tabcolsep}
  \caption{Evaluation on AGQA's more compositional steps.}
  {\begin{tabular}{c c c c } 
  \toprule
  More Compositional Steps    & Binary & Open & All \\
    \midrule
    
    PSAC~\cite{PSAC} & 47.65 & 14.81 & 47.19\\
    HME~\cite{HME} & \textbf{48.09} & 20.98 & \textbf{47.72}\\
    HCRN~\cite{hcrn} & 46.96 & \textbf{23.70} & 46.63 \\
    SHG-VQA & \underline{47.13} & \underline{22.66} & \underline{46.97} \\
    \bottomrule
    \end{tabular}}
    \vspace{-10pt}
    \label{tab:agqa-more-comp-steps}
\end{table}

\begin{table}
  \caption{Evaluation on AGQA's indirect references test metric.}
  \label{tab:agqa-indirectrefs}
  \renewcommand{\arraystretch}{.9}
  \centering \scriptsize \setlength{\tabcolsep}{.25\tabcolsep}
  \begin{tabular}{lccccccccccccc } 
  \toprule
  Method & && \multicolumn{3}{c}{Object} && \multicolumn{3}{c}{Action} && \multicolumn{3}{c}{Temporal} \\
  \cmidrule{4-6} \cmidrule{8-10} \cmidrule{12-14}
  & && B & O & All && B & O & All && B & O & All \\
  \midrule

  \multirow{4}{*}{Precision} & PSAC && 63.69 & 53.77 & 56.64 && 61.01 & 52.46 & 53.24 && 57.52 & 53.74 & 54.39 \\
  & HME && 62.95 & 52.31 & 55.39 && 58.21 & 48.12 & 49.04 && 55.99 & 52.42 & 53.04 \\
  & HCRN && 54.06 & 67.24 & 63.43 && 53.87 & 64.43 & 63.47 && 52.35 & 66.84 & 64.34\\
  & SHG-VQA && \textbf{76.93} & \textbf{86.27} & \textbf{81.59} && \textbf{79.55} & \textbf{87.40} & \textbf{84.90} && \textbf{69.08} & \textbf{87.35} & \textbf{82.87}\\
  
  \midrule 
  
   \multirow{4}{*}{Recall} & PSAC && 45.06 & 27.36 & 38.80 && 40.91 & 22.18 & 25.96 && 35.13 & 26.84 & 30.64\\
   & HME && 46.03 & 26.80 & 39.23 && 41.32 & 21.71 & 25.67 && 37.96 & 26.59 & 31.80\\
   & HCRN && 44.84 & 35.46 & 41.52 && 44.01 & 30.43 & 33.17 && 35.11 & 34.38 & 34.71\\
   & SHG-VQA && \textbf{53.86} & \textbf{43.98} & \textbf{49.85} && \textbf{54.16} & \textbf{37.83} & \textbf{43.10} && \textbf{52.21} & \textbf{43.57} & \textbf{46.78}\\

        \bottomrule
        
      \end{tabular}
      \label{tab:table7}
      \vspace{-10pt}
    \end{table}






\section{Experiments} \label{sec:experiments}
\subsection{Datasets} \label{sec:dataset}
\paragraph{AGQA Benchmark~\cite{grunde2021agqa}:} 
The Action Genome Question Answering benchmark is a visual dataset comprising 192M hand-crafted questions about 9.6K videos from the Charades dataset~\cite{charades}. In addition to VQA accuracy, AGQA presents three testing metrics for testing the VQA methods: indirect references, novel compositions, and more compositional steps.
We use the AGQA 2.0 Balanced dataset, which consists of 2.27M question-answer pairs as a result of balancing the original dataset using stricter procedures to reduce as much language bias as possible. Of the 2.27M questions, there are approximately 1.6M training questions and 669K test questions \cite{AGQA2.0}. To have a standard train-val-test setup for our experiments, we randomly sampled 10\% QA pairs from training data for validation of hyperparameters.
\vspace{-12pt}
\paragraph{STAR Benchmark~\cite{wu2021star}:} 
The STAR dataset provides 60K situated reasoning questions based on 22K trimmed situation video clips, also based on the Charades dataset~\cite{charades}.
They further provide $\sim$144K ground-truth situation graphs including 111 actions, 37 unique objects, and 24 relationships. 
The dataset is split into training, validation, and test sets where test evaluation can be done on the evaluation server a limited number of times. We perform ablations and analysis on the validation set and report test set results to compare with the baselines in Table~\ref{tab:star-main}.

\begin{table*}
  \caption{Results on STAR dataset. \textbf{Best results} are shown in bold font and \underline{second best} results are underlined. Numbers are reported for VQA accuracy in percentages.}
  \label{tab:star-main}
  \renewcommand{\arraystretch}{.9}
  \centering \footnotesize \setlength{\tabcolsep}{.7\tabcolsep}
  \begin{tabular}{lcccccccc}
    \toprule
 
    \multirow{2}{*}{ Method(test)} & Backbone & Obj.  & Hyper.  & \multicolumn{4}{c}{Question Type} \\
    \cmidrule{5-8}
    & & & & Inter. & Seq. & Pred. & Feas. & Overall \\
    \midrule
    Q-type (Random)~\cite{johnson2017clevr} & - & \xmark & \xmark & 25.06 & 24.93 & 24.79 & 24.81 & 24.89\\
    Q-type (Frequent)~\cite{johnson2017clevr} & - & \xmark & \xmark & 19.09 & 19.45 & 12.90 & 18.31 & 17.44\\
    \hline
    Blind Model (LSTM)~\cite{lstm} & GloVe & \xmark & \xmark & 32.24 & 32.17 & 28.56 & 28.41 & 30.34\\
    Blind Model (BERT)~\cite{devlin2018bert} & BERT & \xmark & \xmark & 32.68 & 34.21 & 29.98 & 29.26 & 31.53\\
    \hline
    CNN-LSTM~\cite{yi2019clevrer} & ResNext101-K400 & \xmark & \xmark & 33.25 & 32.67 & 30.69 & 30.43 & 31.76\\
    CNN-BERT~\cite{li2019visualbert} & ResNext101-K400 & \xmark & \xmark & 33.59 & 37.16 & 30.95 & 30.84 & 33.14\\
    \hline
    LCGN~\cite{lcgn} & ResNext101-K400 & \cmark & \xmark & 39.01 & 37.97 & 28.81 & 26.98 & 33.19\\
    HRCN~\cite{hcrn} & ResNext101-K400 & \cmark & \xmark & 39.10 & 38.17 & 28.75 & 27.27 & 33.32\\
    ClipBERT~\cite{clipbert} & ResNext101-K400 & \xmark & \xmark & 39.81 & \textbf{43.59} & 32.24 & \underline{31.42} & 36.70\\
    NS-SR~\cite{wu2021star} & ResNext101-K400 & \cmark & \cmark & 30.88 & 31.76 & 30.23 & 29.73 & 30.65\\
    \hline
    SHG-VQA (Ours) & SlowR50-K400 & \xmark & \cmark &   \textbf{47.98} & 42.03 & \textbf{35.34} & \textbf{32.52} & \textbf{39.47}\\
    SHG-VQA (Ours)& ResNext101-ImageNet1K & \xmark & \cmark & \underline{45.8}  & \underline{42.77} & \underline{34.64} & 29.91 & \underline{38.28}\\
    \bottomrule
  \end{tabular}
  \vspace{-10pt}
\end{table*}

\subsection{Implementations} \label{sec:training}
On the VQA task, we report accuracy; we also report mAP for situation hyper-graph predictions. For AGQA's \textit{indirect references} testing metric, precision and recall are reported. 
For STAR benchmark, we follow the same training protocol as~\cite{wu2021star} and train SHG-VQA from scratch on each question type separately unless specified otherwise.
Training details are shared in the supplementary document. 
\vspace{2pt}

\noindent \textbf{Visual Embeddings:} 
The video frames are resized to size $224\times224$ with a clip length of 16 frames. We use RandAugment~\cite{cubuk2020randaugment} for data augmentation during training. The video clip is input to a  pretrained convolutional network with freeze weights to obtain video features of size $16\times7\times7\times2048$. 
A 2-layer 3D convolutional block with the kernel of size $5\times\ 3 \times 3$ further processes the zero-padded extracted features yielding features of size $8\times7\times7\times d$. The spatio-temporal dimensions are then flattened to obtain a sequence of length $Thw = 392$ $d-$dimensional tokens where $d=768$. The input encoders as well as situation hyper-graph decoders use $L=5$ transformer encoder layers with non-shared weights. \vspace{2pt}
\noindent\textbf{Query embeddings for action and relationship predicates:}
The features output from the video encoder is then input to a situation hyper-graph decoder comprising transformer-based action and relationship decoders. 
Our best model uses $M=8$ relation queries and $N=3$ action queries for each situation; $T$ is set to 16. 
\textbf{AGQA} has 157 total raw actions, 36 unique objects, and 44 unique relationships which obtain 456 relationship triplets $<v_x, e_i, v_y>$ in the training and validation set.
For \textbf{STAR} dataset, there are 37 objects, and 24 relationships yielding 563 unique relation predicates; it also has 111 action classes.
\vspace{2pt}

\noindent \textbf{Situation graph embeddings:}
The decoded situation graph queries are input to a cross-attentional transformer. The input situation graph embedding is a sum of 4 different encoding types: 
1) decoded query embeddings for predicted actions and relationship predicates,
2) situation IDs denoting the situation (or frame) number for each query, 3) attention mask set to 1 for actual tokens and 0 for padded tokens, and 4) token type embeddings to distinguish between action tokens and relationship tokens as shown in Fig.~\ref{fig:graph-embed}. 
Our network uses a 2-layer co-attentional module~\cite{lu2019vilbert} for cross-attention. \\


\noindent \textbf{AGQA:} For AGQA, we train our model with SlowR50 backbone and report results for VQA accuracy on the test set. We also report our model's generalization capability to indirect references. Furthermore, we train our network to report its generalization to novel compositions and to more compositional steps. For more compositional steps, we train our network with randomly sampled 100K QA pairs. 

\noindent \textbf{STAR:} Following~\cite{wu2021star}, we train the model for each question type separately to compare with the baselines. However, this protocol is expensive in terms of time and resources. To address the matter of limited resources, we also tried combining all questions together after the questions filtering from interaction and sequence types and train a single model instead of multiple trainings. We use this training regime to study models ablations.
\vspace{-7pt}
\section{Results and Analysis} \label{sec:results}
\subsection{Comparison to State-of-the-Art}
\noindent \textbf{AGQA:} We compare our method with the existing state-of-the-art methods on AGQA benchmark. The best baseline on AGQA is HCRN~\cite{hcrn} for overall accuracy. HCRN uses appearance features from ResNet101 as well as motion features from ResNext101-Kinetics400 backbones. Our model outperforms HCRN by a significant margin of 7.09\% (HCRN: 42.11\% vs. SHG-VQA: 49.20\%) in terms of overall accuracy (see Table~\ref{tab:agqamain-table}). We observe the biggest improvement of 14.63\% absolute points on object-action reasoning questions compared to the best model in that category i.e., PSAC~\cite{PSAC}: 50.00\% vs. SHG-VQA: 64.63\%. We further report results on the three novel testing metrics as follows: 

\noindent \textbf{a) Novel Compositions:} For novel composition at test time, we observe an overall gain of 4.70\% when compared to the best contender, i.e., PSAC~\cite{PSAC}. For open-ended questions, we outperform HCRN by 6.56\% (see Table~\ref{tab:agqa-novel-comp}); 

\noindent \textbf{b) Indirect References:} When tested for indirect references questions in Table~\ref{tab:agqa-indirectrefs}, we outperform all baselines by an absolute 7.83\%-16.62\% in terms of recall; For precision, we also gain similar improvements;

\noindent \textbf{c) Compositional Steps:} Our model is trained on only 15\% (100K QA pairs) of the training data and is still able to perform on par with the baselines (trained on 1.6M QA pairs) achieving second best results for each category of more compositional steps (table~\ref{tab:agqa-more-comp-steps}). 

\noindent \textbf{STAR:}
For STAR dataset, we first compare the proposed architecture to other state-of-the-art works in the field using SlowR50~\cite{hara2017learning} video backbone and a ResNext101~\cite{Xie_2017_CVPR} as a frame-level backbone  to evaluate accuracy based on 3D video and 2D image architectures. It shows that for both settings, SHG-VQA significantly outperforms other baseline methods even with weaker backbones (see Table~\ref{tab:star-main}).
Concretely, we obtain an absolute gain of $8.53\%$ over NS-SR~\cite{wu2021star}, $5.86\%$ compared to HCRN~\cite{hcrn}, $5.99\%$ improvement over LCGN~\cite{lcgn}, and $\sim2.5\%$ over ClipBERT~\cite{clipbert} which is a SOTA model in terms of overall VQA accuracy. We notice the substantial gain of $7.86\%$ for interaction questions which test the understanding of interactions between entities in a situation. Prediction is the next category of questions that benefits the most with $2.48\%$ improvement.

\subsection{Ablation and Hyperparameter Analysis}
We perform our ablation studies on the STAR benchmark as discussed below:

\noindent \textbf{Impact of situation graphs quality:}
To assess the effect of situation graphs' quality on VQA accuracy, we train a baseline version of our system, where the model is trained on ground truth situation graphs for VQA tasks only (Table ~\ref{tab:hyper-graph-qual}). Since the ground truth situation graphs are not available for the test set, we can only compare SHG-VQA and this baseline on the validation set.  As expected, when taking ground truth situation graphs as input, the performance is significantly improved for interaction (GT=91.9\% vs. predicted=46.78\%) and sequence (GT=80.5\% vs. predicted=42.52\%) questions. However, for the questions about the unseen part of the video, the model with ground truth graph tokens still struggles despite better performance compared to SHG-VQA: prediction (GT=41.22\% vs. predicted=37.82\%) and feasibility (GT=35.42\% vs. predicted=33.61\%). 
We also report the mAP scores for the prediction of action and relationship predicates using our best model in Table~\ref{tab:hg-qual}. We obtain an overall mAP of 87.63 for actions and 72.9 for relationships respectively. \\
\noindent \textbf{Input to cross-attentional transformer:} To evaluate the choice of input to the cross-attentional transformer, we experiment with three settings: a) question and video embeddings, b) question and situation graphs embeddings, c) question, situation graphs, and video embeddings. We observe no gain in the overall VQA accuracy when adding video embeddings to the cross-attentional transformer and get our best results with (b) (table~\ref{tab:cross-attn-inputs}). \\
\noindent \textbf{Situation hyper-graph components:} To study the impact of different components of situation graphs, we train our system with only action predicate tokens with objective $L_{act} + L_{vqa}$, only relationship predicates (with $L_{rel} + L_{vqa}$), and the full model (eq.~\ref{eq:2}). The action predicates are more effective compared to the relationship predicates. When compared in terms of predicate classification, we observe high accuracy for action predicates. Nonetheless, using the full situation graphs perform better than omitting actions or relationship prediction task (table ~\ref{tab:ablations-val}).   \\
\noindent \textbf{Number of queries:} Number of action queries $M$ and relationship queries $N$ is a hyperparameter for SHG-VQA. We report the performance of the SHG-VQA with a varying number of queries for actions and relationships at each timestep in Table~\ref{tab:ablations-val}. We report our best results with $M=3$ and $N=8$.
Additional results are reported in the supplementary document.
\begin{figure}[t]
\begin{center}
  \includegraphics[width=\linewidth]{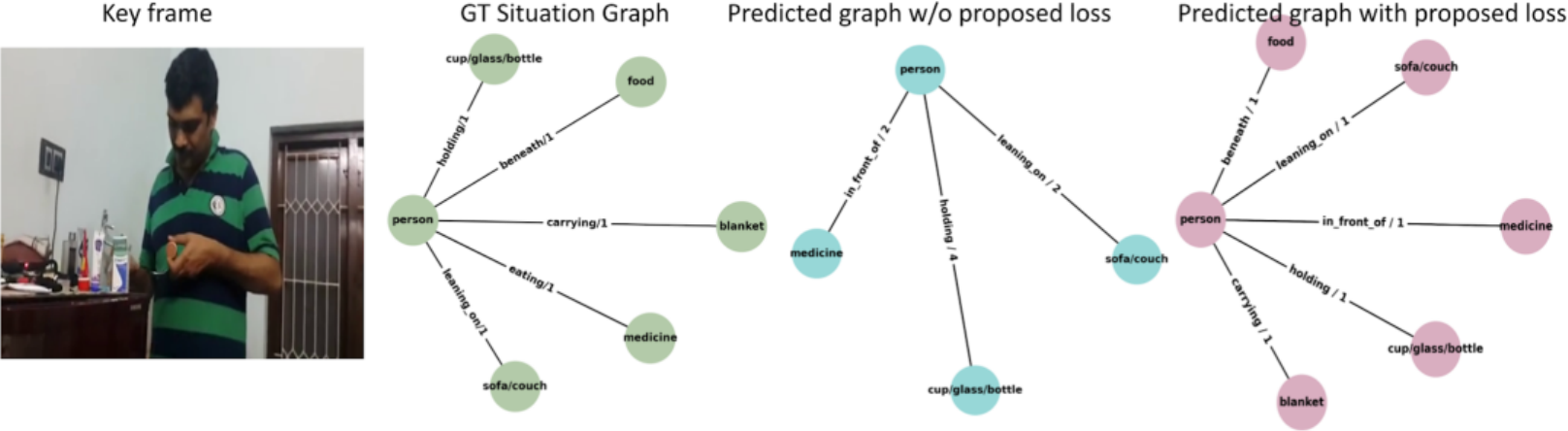}
 \end{center}
 \vspace{-10pt}
  \caption{\small Qualitative example for using frame-wise set prediction loss. Col. 1 shows the frame, col. 2 shows the ground-truth situation graph, col. 3 shows predicted graphs when trained with set prediction loss for the full video, and col. 4 shows the predicted graph when the model is trained by matching each timestep $t$. The edges show the person-object relationship labels along with the number of times it was predicted.} \label{fig:qual}
\end{figure}

\begin{table}
  \caption{\textbf{Impact of hyper-graphs (HG) quality.} Results shown for STAR val set with SlowR50 backbone. }
  \label{tab:hyper-graph-qual}
   \renewcommand{\arraystretch}{.9}
  \centering \scriptsize \setlength{\tabcolsep}{.3\tabcolsep}
  \begin{tabular}{lccccc}
    \toprule
    
     Method    & Interaction & Sequence & Prediction & Feasibility & Overall \\
    \midrule
    Predicted hyper-graphs &  47.08 & 42.52 & 37.82 & 33.61 & 40.26\\
    GT hyper-graphs   &  \textbf{91.9} & \textbf{80.5} & \textbf{41.22} & \textbf{35.42}  & \textbf{62.46}\\
 

    \bottomrule
  \end{tabular}
\end{table}

\begin{table}
  \caption{{\bf Predicate classification results} for situation hyper-graphs in terms of mAP for STAR validation set from SHG-VQA with SlowR50 backbone. Numbers are reported in percentages.}
  \label{tab:hg-qual}
   \renewcommand{\arraystretch}{.9}
  \centering \scriptsize \setlength{\tabcolsep}{.3\tabcolsep}
  \begin{tabular}{lccccc}
    \toprule
 
       &  Interact & Sequence & Prediction & Feasibility & Overall \\
    \midrule
 Actions & 84.77 & 89.43 &  85.30 & 91.46 & 87.63\\
 Relationships & 72.83 & 73.5 & 70.05 & 72.82 & 72.9\\
    \bottomrule
  \end{tabular}
  \vspace{-5pt}
\end{table}

\begin{table}
  \caption{\textbf{Results for cross-attention input.} Results shown for STAR test set with SlowR50 backbone. }
  \label{tab:cross-attn-inputs}
   \renewcommand{\arraystretch}{.9}
  \centering \scriptsize \setlength{\tabcolsep}{.3\tabcolsep}
  \begin{tabular}{lcccccc}
    \toprule
    Method    & Interaction & Sequence & Prediction & Feasibility & Overall \\
    \midrule
    Q + V &  33.28 & 35.60 & 27.93 & 26.43 & 30.81\\
    
    Q + HG &   \textbf{47.98} & 42.03 & \textbf{35.34} & \underline{32.52} & \textbf{39.47}\\
    Q + V + HG &  \underline{45.45} & \textbf{44.19} & \underline{34.22} & \textbf{32.87} & \underline{39.18}\\
 
    \bottomrule
  \end{tabular}
  \vspace{-5pt}
\end{table}

\begin{table}
  \caption{Model variations on STAR validation set with a single model using SlowR50 backbone for all question types. \textbf{Best results} are shown in bold font and \underline{second best} results are underlined. Numbers are reported for VQA accuracy in percentages.}
  \label{tab:ablations-val}
   \renewcommand{\arraystretch}{.9}
  \centering \scriptsize \setlength{\tabcolsep}{.3\tabcolsep}
  \begin{tabular}{lccccc}
    \toprule
 
    Method (val)    &  Interaction & Sequence & Prediction & Feasibility & Overall \\
    \midrule
 
    \underline{\textit{hyper-graph components}} 
    \\
    Action only -- Act=3  &  \underline{40.94} & \underline{38.08} & \underline{35.58} & \underline{30.56} & \underline{38.68}\\
    Relation. only -- Rel=8 &  35.94 & 35.79 & 34.78 & 27.86 & 35.16\\
    Both -- Act=3, Rel=8 &  \textbf{42.93} & \textbf{38.20} & \textbf{36.06} & \textbf{30.56} & \textbf{39.20}\\
    \midrule
    \underline{\textit{Number of queries}} 
    \\
    
    Act=2, Rel=8 &  40.32 &  \underline{40.13} &  \underline{38.78} &   29.73 & 38.34\\
    Act=3, Rel=8 &  \textbf{42.93} &   38.20 & 36.06 &   \underline{30.56} &	\underline{39.20}\\
    Act=4, Rel=8 &  38.40 &   35.79 &  35.58 &  30.35 & 37.06\\
    Act=3, Rel=12 &  \underline{41.32} &  \textbf{40.80} &  \textbf{39.42} &   \textbf{31.39} & \textbf{39.90}\\
    Act=4, Rel=12 &  40.40 &   39.05 &  36.86 &   29.11 & 38.39\\

    \bottomrule
  \end{tabular}
  \vspace{-10pt}
\end{table}

\noindent \textbf{Frame-wise set prediction loss:} Using the notion of time $t$ in the set prediction loss alleviates the problem of duplicate predictions within each situation. See Fig.~\ref{fig:qual} for qualitative examples of produced hyper-graphs with and without this loss. Extra examples are in the supplementary document.\\


\vspace{-15pt}
\section{Conclusion} \label{sec:conclusion}
We presented a novel approach to model situation graph prediction as an underlying sub-task for video question answering. The proposed method predicts a situation hyper-graph structure composed of existing actions and relationships in the input video. The input question can then reason over the predicted graph to solve VQA. We show the impact of the proposed approach by evaluating on two video question-answering benchmarks and achieving significant performance gains overall baseline methods. Our method demonstrates promise for further research in this direction to improve VQA systems even further.  \vspace{5pt}

\noindent \textbf{\footnotesize Acknowledgements.} \footnotesize Aisha Urooj is supported by the ARO grant W911NF-19-1-0356. The U.S. Government is authorized to reproduce and distribute reprints for Governmental purposes notwithstanding any copyright annotation thereon. This work was partially supported under NSF grant CNS-2050731, Research Experience for Undergraduates in Computer Vision.  Dr. Gan is supported by the MIT-IBM Watson AI Lab, DSO grant DSOCO21072, and gift funding from MERL, Cisco, Sony, and Amazon. \textbf{Disclaimer:} The views and conclusions contained herein are those of the authors and should not be interpreted as necessarily representing the official policies or endorsements, either expressed or implied, of ARO, IARPA, DOI/IBC, or the U.S. Government. 

\newpage
{\small
\bibliographystyle{ieee_fullname}
\bibliography{cvpr}

\begin{thebibliography}{10}\itemsep=-1pt

\bibitem{detr}
Nicolas Carion, Francisco Massa, Gabriel Synnaeve, Nicolas Usunier, Alexander
  Kirillov, and Sergey Zagoruyko.
\newblock End-to-end object detection with transformers.
\newblock In {\em European conference on computer vision}, pages 213--229.
  Springer, 2020.

\bibitem{chang2021comprehensive}
Xiaojun Chang, Pengzhen Ren, Pengfei Xu, Zhihui Li, Xiaojiang Chen, and
  Alexander~G Hauptmann.
\newblock A comprehensive survey of scene graphs: Generation and application.
\newblock {\em IEEE Transactions on Pattern Analysis and Machine Intelligence},
  2021.

\bibitem{cubuk2020randaugment}
Ekin~D Cubuk, Barret Zoph, Jonathon Shlens, and Quoc~V Le.
\newblock Randaugment: Practical automated data augmentation with a reduced
  search space.
\newblock In {\em Proceedings of the IEEE/CVF Conference on Computer Vision and
  Pattern Recognition Workshops}, pages 702--703, 2020.

\bibitem{damodaran2021understanding}
Vinay Damodaran, Sharanya Chakravarthy, Akshay Kumar, Anjana Umapathy, Teruko
  Mitamura, Yuta Nakashima, Noa Garcia, and Chenhui Chu.
\newblock Understanding the role of scene graphs in visual question answering.
\newblock {\em arXiv preprint arXiv:2101.05479}, 2021.

\bibitem{dang2021hierarchical}
Long~Hoang Dang, Thao~Minh Le, Vuong Le, and Truyen Tran.
\newblock Hierarchical object-oriented spatio-temporal reasoning for video
  question answering.
\newblock {\em arXiv preprint arXiv:2106.13432}, 2021.

\bibitem{devlin2018bert}
Jacob Devlin, Ming-Wei Chang, Kenton Lee, and Kristina Toutanova.
\newblock Bert: Pre-training of deep bidirectional transformers for language
  understanding.
\newblock {\em arXiv preprint arXiv:1810.04805}, 2018.

\bibitem{dosovitskiy2020image}
Alexey Dosovitskiy, Lucas Beyer, Alexander Kolesnikov, Dirk Weissenborn,
  Xiaohua Zhai, Thomas Unterthiner, Mostafa Dehghani, Matthias Minderer, Georg
  Heigold, Sylvain Gelly, et~al.
\newblock An image is worth 16x16 words: Transformers for image recognition at
  scale.
\newblock {\em arXiv preprint arXiv:2010.11929}, 2020.

\bibitem{fan2019heterogeneous}
Chenyou Fan, Xiaofan Zhang, Shu Zhang, Wensheng Wang, Chi Zhang, and Heng
  Huang.
\newblock Heterogeneous memory enhanced multimodal attention model for video
  question answering.
\newblock In {\em Proceedings of the IEEE/CVF conference on computer vision and
  pattern recognition}, pages 1999--2007, 2019.

\bibitem{HME}
Chenyou Fan, Xiaofan Zhang, Shu Zhang, Wensheng Wang, Chi Zhang, and Heng
  Huang.
\newblock Heterogeneous memory enhanced multimodal attention model for video
  question answering.
\newblock {\em CoRR}, abs/1904.04357, 2019.

\bibitem{gao2018motion}
Jiyang Gao, Runzhou Ge, Kan Chen, and Ram Nevatia.
\newblock Motion-appearance co-memory networks for video question answering.
\newblock In {\em Proceedings of the IEEE Conference on Computer Vision and
  Pattern Recognition}, pages 6576--6585, 2018.

\bibitem{grunde2021agqa}
Madeleine Grunde-McLaughlin, Ranjay Krishna, and Maneesh Agrawala.
\newblock Agqa: A benchmark for compositional spatio-temporal reasoning.
\newblock In {\em Proceedings of the IEEE/CVF Conference on Computer Vision and
  Pattern Recognition}, pages 11287--11297, 2021.

\bibitem{AGQA2.0}
Madeleine Grunde-McLaughlin, Ranjay Krishna, and Maneesh Agrawala.
\newblock Agqa 2.0: An updated benchmark for compositional spatio-temporal
  reasoning, 2022.

\bibitem{hara2017learning}
Kensho Hara, Hirokatsu Kataoka, and Yutaka Satoh.
\newblock Learning spatio-temporal features with 3d residual networks for
  action recognition.
\newblock In {\em Proceedings of the IEEE International Conference on Computer
  Vision Workshops}, pages 3154--3160, 2017.

\bibitem{lstm}
Sepp Hochreiter and J{\"u}rgen Schmidhuber.
\newblock Long short-term memory.
\newblock {\em Neural computation}, 9(8):1735--1780, 1997.

\bibitem{lcgn}
Ronghang Hu, Anna Rohrbach, Trevor Darrell, and Kate Saenko.
\newblock Language-conditioned graph networks for relational reasoning.
\newblock In {\em Proceedings of the IEEE International Conference on Computer
  Vision}, pages 10294--10303, 2019.

\bibitem{jang2019video}
Yunseok Jang, Yale Song, Chris~Dongjoo Kim, Youngjae Yu, Youngjin Kim, and
  Gunhee Kim.
\newblock Video question answering with spatio-temporal reasoning.
\newblock {\em International Journal of Computer Vision}, 127(10):1385--1412,
  2019.

\bibitem{jang2017tgif}
Yunseok Jang, Yale Song, Youngjae Yu, Youngjin Kim, and Gunhee Kim.
\newblock Tgif-qa: Toward spatio-temporal reasoning in visual question
  answering.
\newblock In {\em Proceedings of the IEEE Conference on Computer Vision and
  Pattern Recognition}, pages 2758--2766, 2017.

\bibitem{jiang2020divide}
Jianwen Jiang, Ziqiang Chen, Haojie Lin, Xibin Zhao, and Yue Gao.
\newblock Divide and conquer: Question-guided spatio-temporal contextual
  attention for video question answering.
\newblock In {\em Proceedings of the AAAI Conference on Artificial
  Intelligence}, pages 11101--11108, 2020.

\bibitem{jiang2020reasoning}
Pin Jiang and Yahong Han.
\newblock Reasoning with heterogeneous graph alignment for video question
  answering.
\newblock In {\em Proceedings of the AAAI Conference on Artificial
  Intelligence}, pages 11109--11116, 2020.

\bibitem{xnm}
Juanzi~Li Jiaxin~Shi, Hanwang~Zhang.
\newblock Explainable and explicit visual reasoning over scene graphs.
\newblock In {\em CVPR}, 2019.

\bibitem{johnson2017clevr}
Justin Johnson, Bharath Hariharan, Laurens van~der Maaten, Li Fei-Fei,
  C~Lawrence Zitnick, and Ross Girshick.
\newblock Clevr: A diagnostic dataset for compositional language and elementary
  visual reasoning.
\newblock In {\em CVPR}, 2017.

\bibitem{kant2020spatially}
Yash Kant, Dhruv Batra, Peter Anderson, Alexander Schwing, Devi Parikh, Jiasen
  Lu, and Harsh Agrawal.
\newblock Spatially aware multimodal transformers for textvqa.
\newblock In {\em Computer Vision--ECCV 2020: 16th European Conference,
  Glasgow, UK, August 23--28, 2020, Proceedings, Part IX 16}, pages 715--732.
  Springer, 2020.

\bibitem{khan2021reason}
Aisha~Urooj Khan, Hilde Kuehne, Kevin Duarte, Chuang Gan, Niels Lobo, and
  Mubarak Shah.
\newblock Found a reason for me? weakly-supervised grounded visual question
  answering using capsules, 2021.

\bibitem{khan2022_wsg_vlt}
Aisha~Urooj Khan, Hilde Kuehne, Chuang Gan, Niels Da~Vitoria Lobo, and Mubarak
  Shah.
\newblock Weakly supervised grounding for vqa in vision-language transformers.
\newblock In Shai Avidan, Gabriel Brostow, Moustapha Ciss{\'e}, Giovanni~Maria
  Farinella, and Tal Hassner, editors, {\em Computer Vision -- ECCV 2022},
  pages 652--670, Cham, 2022. Springer Nature Switzerland.

\bibitem{khan2020mmftbert}
Aisha~Urooj Khan, Amir Mazaheri, Niels da Vitoria~Lobo, and Mubarak Shah.
\newblock Mmft-bert: Multimodal fusion transformer with bert encodings for
  visual question answering, 2020.

\bibitem{kim2022pure}
Jinwoo Kim, Tien~Dat Nguyen, Seonwoo Min, Sungjun Cho, Moontae Lee, Honglak
  Lee, and Seunghoon Hong.
\newblock Pure transformers are powerful graph learners.
\newblock {\em arXiv preprint arXiv:2207.02505}, 2022.

\bibitem{hcrn}
Thao~Minh Le, Vuong Le, Svetha Venkatesh, and Truyen Tran.
\newblock Hierarchical conditional relation networks for video question
  answering.
\newblock {\em CoRR}, abs/2002.10698, 2020.

\bibitem{clipbert}
Jie Lei, Linjie Li, Luowei Zhou, Zhe Gan, Tamara~L Berg, Mohit Bansal, and
  Jingjing Liu.
\newblock Less is more: Clipbert for video-and-language learning via sparse
  sampling.
\newblock In {\em Proceedings of the IEEE/CVF Conference on Computer Vision and
  Pattern Recognition}, pages 7331--7341, 2021.

\bibitem{lei2018tvqa}
Jie Lei, Licheng Yu, Mohit Bansal, and Tamara~L Berg.
\newblock Tvqa: Localized, compositional video question answering.
\newblock In {\em EMNLP}, 2018.

\bibitem{ALBEF}
Junnan Li, Ramprasaath~R. Selvaraju, Akhilesh~Deepak Gotmare, Shafiq Joty,
  Caiming Xiong, and Steven Hoi.
\newblock Align before fuse: Vision and language representation learning with
  momentum distillation.
\newblock In {\em NeurIPS}, 2021.

\bibitem{li2019relation}
Linjie Li, Zhe Gan, Yu Cheng, and Jingjing Liu.
\newblock Relation-aware graph attention network for visual question answering.
\newblock In {\em Proceedings of the IEEE/CVF international conference on
  computer vision}, pages 10313--10322, 2019.

\bibitem{li2019visualbert}
Liunian~Harold Li, Mark Yatskar, Da Yin, Cho-Jui Hsieh, and Kai-Wei Chang.
\newblock Visualbert: A simple and performant baseline for vision and language.
\newblock {\em arXiv preprint arXiv:1908.03557}, 2019.

\bibitem{li2019beyond}
Xiangpeng Li, Jingkuan Song, Lianli Gao, Xianglong Liu, Wenbing Huang, Xiangnan
  He, and Chuang Gan.
\newblock Beyond rnns: Positional self-attention with co-attention for video
  question answering.
\newblock In {\em Proceedings of the AAAI Conference on Artificial
  Intelligence}, pages 8658--8665, 2019.

\bibitem{PSAC}
Xiangpeng Li, Jingkuan Song, Lianli Gao, Xianglong Liu, Wenbing Huang, Xiangnan
  He, and Chuang Gan.
\newblock Beyond rnns: Positional self-attention with co-attention for video
  question answering.
\newblock In {\em AAAI}, 2019.

\bibitem{liu2021hair}
Fei Liu, Jing Liu, Weining Wang, and Hanqing Lu.
\newblock Hair: Hierarchical visual-semantic relational reasoning for video
  question answering.
\newblock In {\em Proceedings of the IEEE/CVF International Conference on
  Computer Vision}, pages 1698--1707, 2021.

\bibitem{liu2021video}
Ze Liu, Jia Ning, Yue Cao, Yixuan Wei, Zheng Zhang, Stephen Lin, and Han Hu.
\newblock Video swin transformer.
\newblock {\em arXiv preprint arXiv:2106.13230}, 2021.

\bibitem{lu2019vilbert}
Jiasen Lu, Dhruv Batra, Devi Parikh, and Stefan Lee.
\newblock Vilbert: Pretraining task-agnostic visiolinguistic representations
  for vision-and-language tasks.
\newblock {\em arXiv preprint arXiv:1908.02265}, 2019.

\bibitem{maharaj2017dataset}
Tegan Maharaj, Nicolas Ballas, Anna Rohrbach, Aaron Courville, and Christopher
  Pal.
\newblock A dataset and exploration of models for understanding video data
  through fill-in-the-blank question-answering.
\newblock In {\em Proceedings of the IEEE Conference on Computer Vision and
  Pattern Recognition}, pages 6884--6893, 2017.

\bibitem{mao2018neuro}
Jiayuan Mao, Chuang Gan, Pushmeet Kohli, Joshua~B Tenenbaum, and Jiajun Wu.
\newblock The neuro-symbolic concept learner: Interpreting scenes, words, and
  sentences from natural supervision.
\newblock In {\em International Conference on Learning Representations}, 2018.

\bibitem{marino2021krisp}
Kenneth Marino, Xinlei Chen, Devi Parikh, Abhinav Gupta, and Marcus Rohrbach.
\newblock Krisp: Integrating implicit and symbolic knowledge for open-domain
  knowledge-based vqa.
\newblock In {\em Proceedings of the IEEE/CVF Conference on Computer Vision and
  Pattern Recognition}, pages 14111--14121, 2021.

\bibitem{patel2021recent}
Devshree Patel, Ratnam Parikh, and Yesha Shastri.
\newblock Recent advances in video question answering: A review of datasets and
  methods.
\newblock In {\em International Conference on Pattern Recognition}, pages
  339--356. Springer, 2021.

\bibitem{peng2021progressive}
Liang Peng, Shuangji Yang, Yi Bin, and Guoqing Wang.
\newblock Progressive graph attention network for video question answering.
\newblock In {\em Proceedings of the 29th ACM International Conference on
  Multimedia}, pages 2871--2879, 2021.

\bibitem{qian2022scene}
Tianwen Qian, Jingjing Chen, Shaoxiang Chen, Bo Wu, and Yu-Gang Jiang.
\newblock Scene graph refinement network for visual question answering.
\newblock {\em IEEE Transactions on Multimedia}, 2022.

\bibitem{ren2015faster}
Shaoqing Ren, Kaiming He, Ross Girshick, and Jian Sun.
\newblock Faster r-cnn: Towards real-time object detection with region proposal
  networks.
\newblock {\em Advances in neural information processing systems}, 28, 2015.

\bibitem{seo-etal-2021-attend}
Ahjeong Seo, Gi-Cheon Kang, Joonhan Park, and Byoung-Tak Zhang.
\newblock Attend what you need: Motion-appearance synergistic networks for
  video question answering.
\newblock In {\em Proceedings of the 59th Annual Meeting of the Association for
  Computational Linguistics and the 11th International Joint Conference on
  Natural Language Processing (Volume 1: Long Papers)}, pages 6167--6177,
  Online, Aug. 2021. Association for Computational Linguistics.

\bibitem{seo2021look}
Paul~Hongsuck Seo, Arsha Nagrani, and Cordelia Schmid.
\newblock Look before you speak: Visually contextualized utterances.
\newblock In {\em Proceedings of the IEEE/CVF Conference on Computer Vision and
  Pattern Recognition}, pages 16877--16887, 2021.

\bibitem{charades}
Gunnar~A Sigurdsson, G{\"u}l Varol, Xiaolong Wang, Ali Farhadi, Ivan Laptev,
  and Abhinav Gupta.
\newblock Hollywood in homes: Crowdsourcing data collection for activity
  understanding.
\newblock In {\em European Conference on Computer Vision}, pages 510--526.
  Springer, 2016.

\bibitem{Sigurdsson2016HollywoodIH}
Gunnar~A. Sigurdsson, G{\"u}l Varol, X. Wang, Ali Farhadi, Ivan Laptev, and
  Abhinav~Kumar Gupta.
\newblock Hollywood in homes: Crowdsourcing data collection for activity
  understanding.
\newblock {\em ECCV}, 2016.

\bibitem{singh2019strings}
Ajeet~Kumar Singh, Anand Mishra, Shashank Shekhar, and Anirban Chakraborty.
\newblock From strings to things: Knowledge-enabled vqa model that can read and
  reason.
\newblock In {\em Proceedings of the IEEE/CVF International Conference on
  Computer Vision}, pages 4602--4612, 2019.

\bibitem{tapaswi2016movieqa}
Makarand Tapaswi, Yukun Zhu, Rainer Stiefelhagen, Antonio Torralba, Raquel
  Urtasun, and Sanja Fidler.
\newblock Movieqa: Understanding stories in movies through question-answering.
\newblock In {\em Proceedings of the IEEE conference on computer vision and
  pattern recognition}, pages 4631--4640, 2016.

\bibitem{vaswani2017attention}
Ashish Vaswani, Noam Shazeer, Niki Parmar, Jakob Uszkoreit, Llion Jones,
  Aidan~N Gomez, {\L}ukasz Kaiser, and Illia Polosukhin.
\newblock Attention is all you need.
\newblock In {\em Advances in neural information processing systems}, pages
  5998--6008, 2017.

\bibitem{wu2021star}
Bo Wu, Shoubin Yu, Zhenfang Chen, Joshua~B Tenenbaum, and Chuang Gan.
\newblock Star: A benchmark for situated reasoning in real-world videos.
\newblock In {\em Thirty-fifth Conference on Neural Information Processing
  Systems}, 2021.

\bibitem{xiao2021next}
Junbin Xiao, Xindi Shang, Angela Yao, and Tat-Seng Chua.
\newblock Next-qa: Next phase of question-answering to explaining temporal
  actions.
\newblock In {\em Proceedings of the IEEE/CVF Conference on Computer Vision and
  Pattern Recognition}, pages 9777--9786, 2021.

\bibitem{xiao2021video}
Junbin Xiao, Angela Yao, Zhiyuan Liu, Yicong Li, Wei Ji, and Tat-Seng Chua.
\newblock Video as conditional graph hierarchy for multi-granular question
  answering.
\newblock {\em arXiv preprint arXiv:2112.06197}, 2021.

\bibitem{Xie_2017_CVPR}
Saining Xie, Ross Girshick, Piotr Dollar, Zhuowen Tu, and Kaiming He.
\newblock Aggregated residual transformations for deep neural networks.
\newblock In {\em Proceedings of the IEEE Conference on Computer Vision and
  Pattern Recognition (CVPR)}, July 2017.

\bibitem{xu2017video}
Dejing Xu, Zhou Zhao, Jun Xiao, Fei Wu, Hanwang Zhang, Xiangnan He, and Yueting
  Zhuang.
\newblock Video question answering via gradually refined attention over
  appearance and motion.
\newblock In {\em Proceedings of the 25th ACM international conference on
  Multimedia}, pages 1645--1653, 2017.

\bibitem{yang2021just}
Antoine Yang, Antoine Miech, Josef Sivic, Ivan Laptev, and Cordelia Schmid.
\newblock Just ask: Learning to answer questions from millions of narrated
  videos.
\newblock In {\em Proceedings of the IEEE/CVF International Conference on
  Computer Vision}, pages 1686--1697, 2021.

\bibitem{yi2019clevrer}
Kexin Yi, Chuang Gan, Yunzhu Li, Pushmeet Kohli, Jiajun Wu, Antonio Torralba,
  and Joshua~B Tenenbaum.
\newblock Clevrer: Collision events for video representation and reasoning.
\newblock {\em ICLR}, 2020.

\bibitem{nsvqa}
Kexin Yi, Jiajun Wu, Chuang Gan, Antonio Torralba, Pushmeet Kohli, and Joshua~B
  Tenenbaum.
\newblock {Neural-Symbolic VQA: Disentangling Reasoning from Vision and
  Language Understanding}.
\newblock In {\em Advances in Neural Information Processing Systems (NIPS)},
  2018.

\bibitem{yu2020ernie}
Fei Yu, Jiji Tang, Weichong Yin, Yu Sun, Hao Tian, Hua Wu, and Haifeng Wang.
\newblock Ernie-vil: Knowledge enhanced vision-language representations through
  scene graph.
\newblock {\em arXiv preprint arXiv:2006.16934}, 2020.

\bibitem{yu2021learning}
Weijiang Yu, Haoteng Zheng, Mengfei Li, Lei Ji, Lijun Wu, Nong Xiao, and Nan
  Duan.
\newblock Learning from inside: Self-driven siamese sampling and reasoning for
  video question answering.
\newblock {\em Advances in Neural Information Processing Systems}, 34, 2021.

\bibitem{yu2019activitynet}
Zhou Yu, Dejing Xu, Jun Yu, Ting Yu, Zhou Zhao, Yueting Zhuang, and Dacheng
  Tao.
\newblock Activitynet-qa: A dataset for understanding complex web videos via
  question answering.
\newblock In {\em Proceedings of the AAAI Conference on Artificial
  Intelligence}, pages 9127--9134, 2019.

\bibitem{zadeh2019social}
Amir Zadeh, Michael Chan, Paul~Pu Liang, Edmund Tong, and Louis-Philippe
  Morency.
\newblock Social-iq: A question answering benchmark for artificial social
  intelligence.
\newblock In {\em Proceedings of the IEEE/CVF Conference on Computer Vision and
  Pattern Recognition}, pages 8807--8817, 2019.

\bibitem{zellers2022merlot}
Rowan Zellers, Jiasen Lu, Ximing Lu, Youngjae Yu, Yanpeng Zhao, Mohammadreza
  Salehi, Aditya Kusupati, Jack Hessel, Ali Farhadi, and Yejin Choi.
\newblock Merlot reserve: Neural script knowledge through vision and language
  and sound.
\newblock {\em arXiv preprint arXiv:2201.02639}, 2022.

\bibitem{zellers2021merlot}
Rowan Zellers, Ximing Lu, Jack Hessel, Youngjae Yu, Jae~Sung Park, Jize Cao,
  Ali Farhadi, and Yejin Choi.
\newblock Merlot: Multimodal neural script knowledge models.
\newblock {\em Advances in Neural Information Processing Systems}, 34, 2021.

\bibitem{DBLP:journals/corr/ZengCCLNS16}
Kuo{-}Hao Zeng, Tseng{-}Hung Chen, Ching{-}Yao Chuang, Yuan{-}Hong Liao,
  Juan~Carlos Niebles, and Min Sun.
\newblock Leveraging video descriptions to learn video question answering.
\newblock {\em CoRR}, abs/1611.04021, 2016.

\bibitem{zhong2022video}
Yaoyao Zhong, Wei Ji, Junbin Xiao, Yicong Li, Weihong Deng, and Tat-Seng Chua.
\newblock Video question answering: Datasets, algorithms and challenges, 2022.

\bibitem{Zhu2022SceneGraphGenerationSurvey}
Guangming Zhu, Liang Zhang, Youliang Jiang, Yixuan Dang, Haoran Hou, Peiyi
  Shen, Mingtao Feng, Xia Zhao, Qiguang Miao, Syed Afaq~Ali Shah, and Mohammed
  Bennamoun.
\newblock Scene graph generation: {A} comprehensive survey.
\newblock {\em CoRR}, abs/2201.00443, 2022.

\bibitem{zhu2017uncovering}
Linchao Zhu, Zhongwen Xu, Yi Yang, and Alexander~G Hauptmann.
\newblock Uncovering the temporal context for video question answering.
\newblock {\em International Journal of Computer Vision}, 124(3):409--421,
  2017.

\end{thebibliography}
}

\appendix

\newpage

\section{ Supplementary Material: Learning to Predict Situation Hyper-Graphs for Video Question Answering}


\noindent In this supplementary document, we discuss the following:
\begin{enumerate}
    \item Additional architectural details (\ref{arch_details})
    \item Implementation and training details (\ref{impl_details})
    \item Additional experimental details (\ref{exp_details})
    \item Additional results and analyses (\ref{quant_results})
    \item Qualitative results (\ref{sec:qual_results-supp})
    \item Computational cost of SHG-VQA (\ref{sec:computational-cost})
    \item Ethical Considerations (\ref{sec:limitations})
        
\end{enumerate}

\section{Additional Architectural Details} \label{arch_details}
In this section, we provide the additional architectural details as follows:

\subsection{Input processing} \label{sec:problem-definition1}
The SHG-VQA can be trained in both open-ended as well as multiple choices settings. For multiple choice setup, $C$ answer choices are also given as input with the question. The goal hence becomes a $C$-way classification task. 

\subsection{Question Encoder:} \label{sec:question-enc1}
We encode the question (and answer choices) as follows:
first, a learnable embedding layer is used to initialize each word token with an embedding vector. These word embeddings along with the special token $[CLS]$ are input to the text transformer encoder encoding each word using multi-head self-attention between different words at each encoder layer. 
In the multiple choices setup, we append the answer choices to the question words as follows: 

\begin{equation} \label{eq:1}
\begin{split}
QA = [CLS] + Q + [SEP] + A0 + [SEP] + A1 \\
+ ...+ [SEP] + A_C
\end{split}
\end{equation}

where, QA is the sequence composed of question words and answer choices separated by a special token $[SEP]$; $[CLS]$ is a special token appended at index $0$ to aggregate question and answer choices into a sentence vector; C denotes the number of answer choices. We empirically find that composing the question and answer choices in the above format gives better performance. 

\subsection{Action Decoder} \label{sec:action-decoder} 
Given the video clip features, we want to decode the set of actions $A$ in each frame. 
To decode the set of actions $A_t$ at each timestep $t$, a sequence of learnable embeddings of size $d$ referred to as action queries of length $|N| \times T$ is input to the action decoder. The action decoder comprises the standard transformer decoder architecture and stacks $L$ decoder layers. In addition, the action decoder also takes encoded video tokens as memory and a target mask (of size $\mathbb{R}^{(|N| \times T) \times (|N| \times T)}$). The target mask is created to perform parallel decoding predicting all actions in a frame at once based on the decoded actions in previous frames. The target mask prevents attending to the action queries from future frames (by setting values to $-\infty$) which masks them out \cite{vaswani2017attention}. Our approach deviates from using traditional approach~\cite{vaswani2017attention} or the parallel decoding approach~\cite{detr} by using a masking at frame-level e.g., to decode actions for frame 0, we set next $|N| \times (T-1)$ positions to $-\infty$ and so on. The action decoder outputs the decoded sequence of action features for each time step t. 

\begin{figure}[t]
\begin{center}

\includegraphics[width=\linewidth]{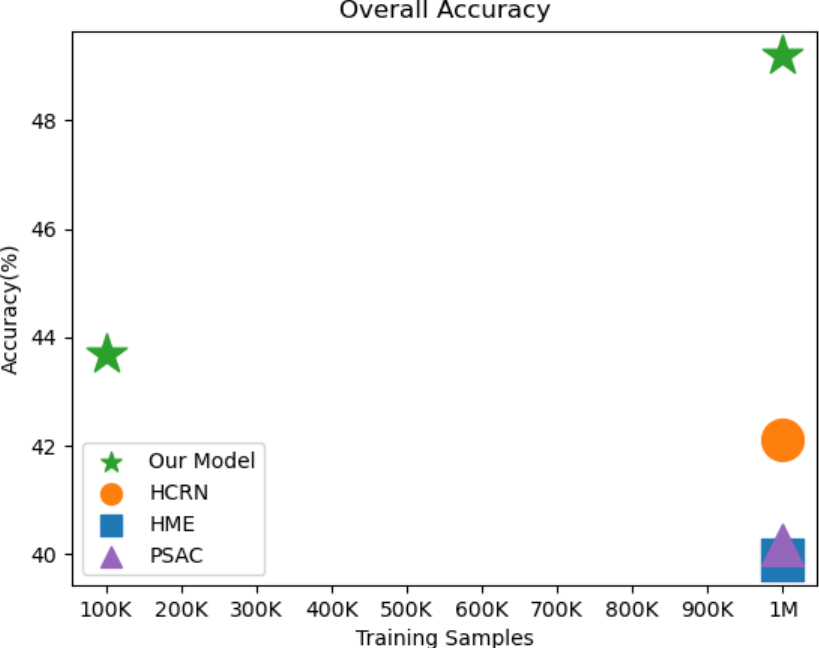}
 
 \end{center}
  \caption{Performance comparison with the baselines w.r.t. training data samples on AGQA dataset. SHG-VQA outperforms the baselines even when trained with only 100K samples.} \label{fig:perform-comp}
\end{figure}

\subsection{Relationship Decoder} \label{sec:relation-decoder}
Like the action decoder, we employ a relationship decoder to decode the set of relations $R_t$ in each frame. We input a sequence of $|M| \times T$ learnable relation embeddings of size $d$ called relation queries and encoded video tokens $x_{V_e}$ as input to the relationship decoder. The relationship decoder has the same architecture as action decoder with variable weights. The relation decoder also takes a target mask (of size $\mathbb{R}^{(|M| \times T) \times (|M| \times T)}$) and perform parallel decoding to predict all relations in a frame at once. The relationships decoder outputs the decoded sequence of relation features for the full video. 

\subsection{Prediction Heads:}  
\begin{figure}[t]
\begin{center}

\includegraphics[width=0.7\linewidth]{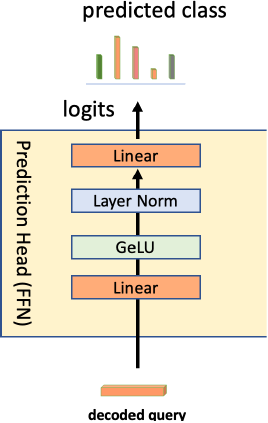}
 
 \end{center}
  \caption{Prediction Head: Each decoded query is passed through a FFN to predict one of the classes or a ``no-class" label.} \label{fig:pred_head}
\end{figure}

Prediction heads take the decoded queries as input and classify them as an action/relationship from the actual classes or the "no-class" (denoted by $\phi$). Therefore, for each prediction head, the total number of classes are $\#classes+1$.
See fig.~\ref{fig:pred_head} for illustration of prediction head.

\begin{figure}[t]
\begin{center}
  \includegraphics[width=0.9\linewidth]{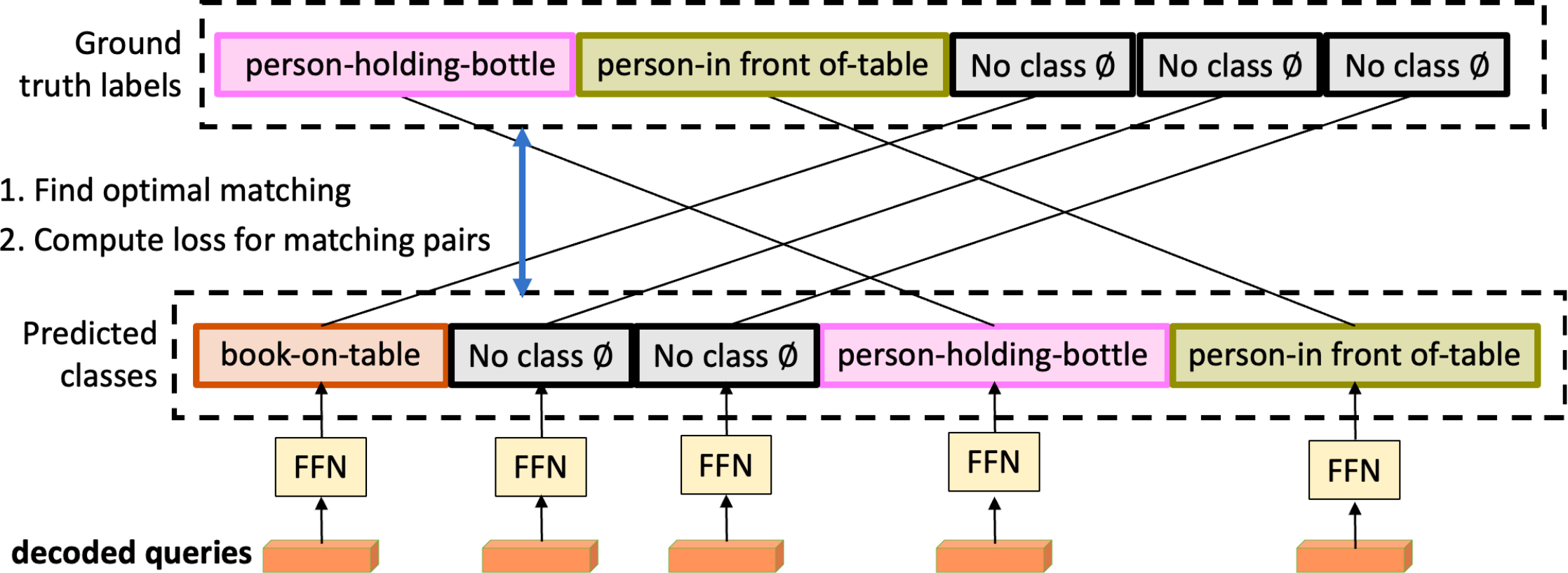}
 \end{center}
  \caption{Overview of the bi-partite matching in SHG-VQA. Optimal bipartite matching ( $\mathcal{L}_{match}(.)$) is performed between the set of predicted classes for all decoded queries (of actions/relationship predicates) and the ground truth labels using the Hungarian algorithm. Per frame optimal matching is carried $\forall t \in \{1,...,T\}$. Then, a loss is computed between the matched pairs of ground truth labels and predicted classes using a cross-entropy loss function. See section 3.4 (main paper) for details.
 } \label{fig:pred_heads}
  \vspace{-10pt}
\end{figure}



\begin{figure*}
\centering
\begin{subfigure}{.3\textwidth}
\includegraphics[width=\linewidth]{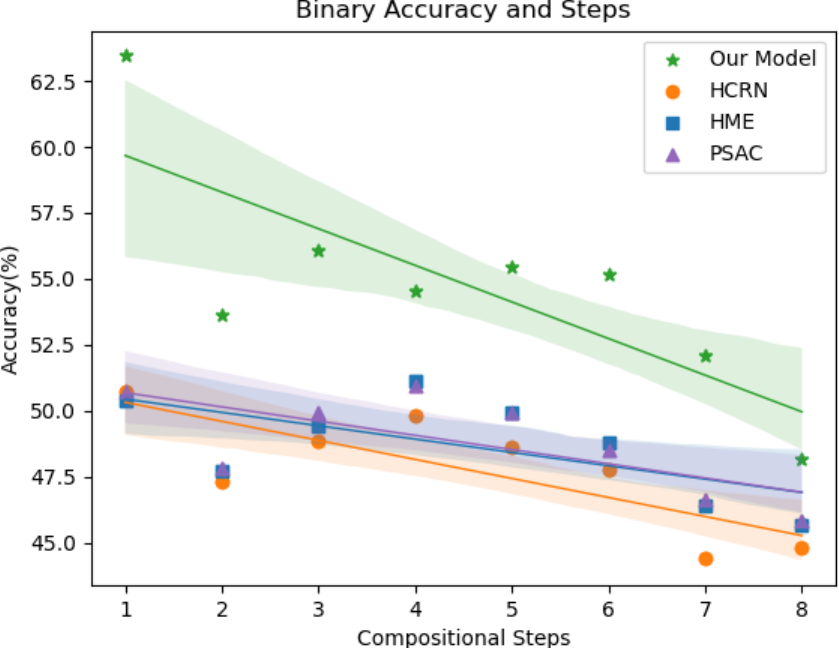}
\end{subfigure}%
\hspace{0.3em}
\begin{subfigure}{.3\textwidth}
\includegraphics[width=\linewidth]{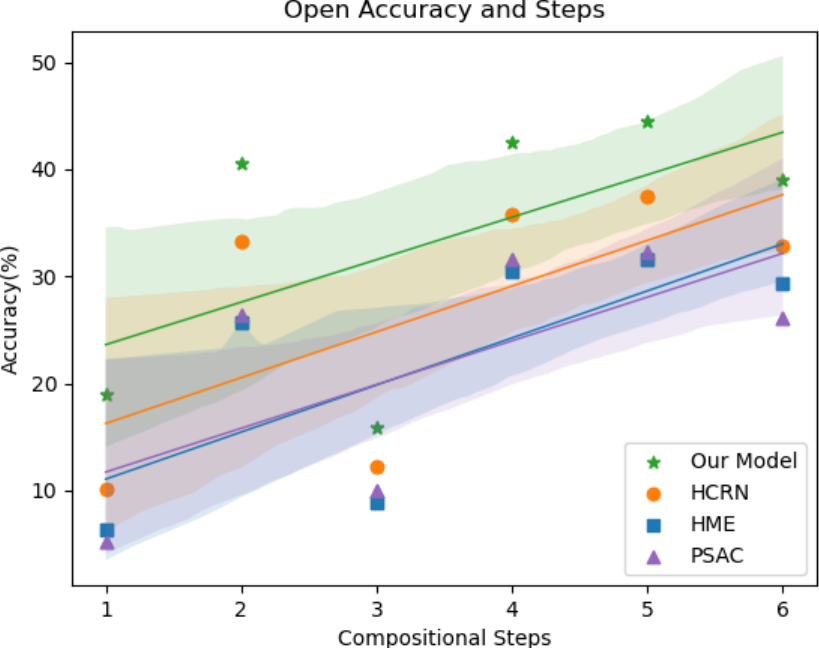}
\end{subfigure}
\begin{subfigure}{.3\textwidth}
\includegraphics[width=\linewidth]{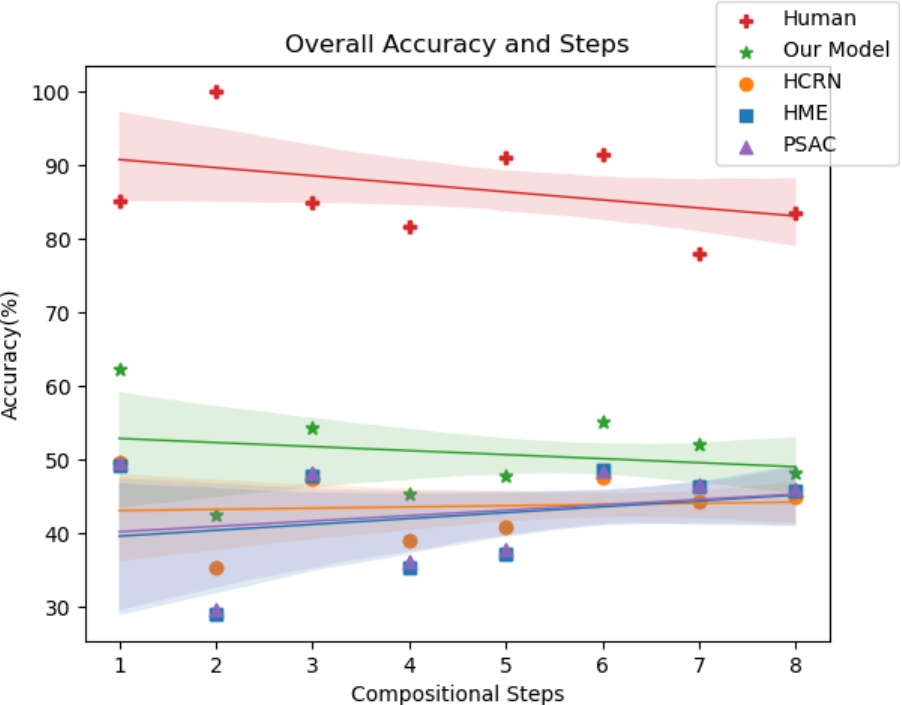}
\end{subfigure}
\caption{Correlation between accuracy and compositional steps for binary answers, open answers, and overall. To do so, a linear regression model is fit for each model's performance. Our model is superior in performance than the baselines bridging the accuracy gap narrower with the human performance. The shaded area indicates 80\% confidence interval.  }
\label{lin_reg_graphs}
\end{figure*}

\section{Additional Implementation Details} \label{impl_details}
\subsection{Training details} SHG-VQA is trained with learning rate $lr=1e-5$, BERT~\cite{devlin2018bert} optimizer, and batch size upto 128 for each training depending on the maximum samples which could fit to the GPUs. The best reported results for both datasets use M=8 relations and N=3 actions.
\subsection{Hypergraph token handling at test time:} During the training of the situation hyper-graph embedding, we use attention mask for masking padded tokens. However, these masks are not available at inference time because we assume that we are only provided the video and question with answer choices at test time. Thus, we set attention mask to all 1's at inference time. Moreover, the prediction is made to the original video clip with out any data augmentation. 

\subsection{Data Augmentations}
\noindent While we obtain a good increase in performance over existing methods for interaction and sequence questions, SHG-VQA performed on par with the baselines for prediction and feasibility. 
We attribute this to the less training data available for prediction and feasibility. 
The number of questions for each question type in training set are: interaction:$\sim$16K , sequence: $\sim$22K, prediction: $\sim$4K, feasibility: $\sim$3K.
To address this matter, we add training samples from other question types. 
We filter out the videos from interaction and sequence question types which have the same video ID as prediction and feasibility. 
This avoids the data leakage problem of peeping into future frames during training. The remaining set of QA pairs are added to the training set for prediction and feasibility resulting in $\sim$15K training samples for feasibility and $\sim$16K samples for prediction question type. We observe the expected performance gain when the model is provided proportional amount of data.

\section{Additional Experiment Details} \label{exp_details}

\subsection{Details about AGQA baselines}
For AGQA, we consider three video QA methods as our baselines: PSAC~\cite{PSAC}, HME~\cite{HME}, and HCRN~\cite{hcrn}. PSAC uses ResNet-152 to extract video features; HME uses ResNet or VGG for appearance features and use C3D for motion features extraction; HCRN uses ResNet101 for appearance features and ResNext101 pretrained on Kinetics-400 to extract motion features. We use SlowR50 as our backbone model.
\subsection{Training Details for STAR Dataset} \label{train-protocol}

\noindent \textbf{Data preprocessing for ablation studies:} Baselines on STAR dataset train a separate model for each question type. Because training separate models for each question time is not feasible in terms of time and computational resources, we merged the data from all question types for our ablations. To do so, we carefully removed the videos and the corresponding QA pairs from interaction and sequence question types which appear in prediction and feasibility questions. As prediction and feasibility questions are about the future frames not available at inference time, keeping these questions for other question types could give an advantage to the model of looking at the full video even if it happens for solving a different question. Filtering out those QA pairs before merging all questions makes it a fair training for prediction and feasibility questions. However, these questions comprises a large chunk for interaction and sequence. As expected, this declines the VQA performance for interaction and sequence questions upto 2\%-8\%. However, we notice a gain over prediction and feasibility questions just by showing more examples to the model even if they are not for the same question types. More specifically, on test set, we notice prediction accuracy of 37.29\% (merged data) vs. 35.34\%(separate) and feasibility accuracy of 33.04\% (merged data) vs. 32.52\%(separate). We also experimented with using questions from all question types without any filtering and obtained the overall validation accuracy of 48.25\%.
In Table~\ref{tab:training-details}, we provide further details about the experiments including batch sizes for each model, backbone, and loss function. All models were trained up to 100 epochs using early stopping based on the validation accuracy.
If not stated otherwise, all ablations are performed with a single model (batch size=32) trained on all questions together with filtering out the QA pairs with overlapped video IDs between \{interaction, sequence\} and \{feasibility, prediction\}.

\noindent \textbf{Different batch size per question type/model:} As each question type is trained on a separate model with different constrains such as amount of data, but constant hardware requirements, we first evaluate different batch sizes for training each model depending on the maximum number of samples could be used for training.
Column 3-\textit{Batch Size} in Table~\ref{tab:training-details} shows a tuple with batch sizes for feasibility, prediction, sequence, and interaction respectively. For training a separate model on each question type, we used batch size=16 with Slow\_R50. For ResNext101, we used batch sizes (16, 16, 4, 4) for (feasibility, prediction, sequence, interaction). 
In our experiments, we observe no significant difference in VQA accuracy when training the models with different batch sizes.  Nonetheless, our best results are reported using batch sizes of (16, 16, 16, 16) with Slow\_R50 backbone, and batch sizes (16, 16, 4, 4) for ResNext101. 

\begin{table*}
  \caption{Results on AGQA dataset for different question types w.r.t vision (\textbf{w)} and question-only (\textbf{w/o}) variants of all models. Best results are shown in \textbf{bold} font, second best results are in \textcolor{blue}{blue} font. SHG-VQA performs better or on par to the baselines with only 100K samples (baselines use 1.6M training samples). Numbers are reported in percentages.
  }
  \label{tab:agqamain-table-supp}
  \renewcommand{\arraystretch}{.9}
  \centering \scriptsize \setlength{\tabcolsep}{.23\tabcolsep}
  \begin{tabular}{lccccccccccccccccccccccc}
    \toprule
&&\multicolumn{8}{c}{Reasoning} & & \multicolumn{3}{c}{Semantic} && \multicolumn{5}{c}{Structure} && \multicolumn{3}{c}{Overall} \\
\cmidrule{3-10} \cmidrule{12-14} \cmidrule{16-20}  \cmidrule{22-24}
Method && obj-rel & rel-action & obj-action & superlative & sequencing & exists & duration & activity && obj & rel & action && query & compare & choose & logic & verify && binary & open & all \\
\midrule
\multirow{2}{*}{PSAC~\cite{PSAC}} & w/o &  37.91 & 49.95 &  50.01 & 33.59 & 49.78 & 50.04 & 45.77 & 4.88 && 38.03 & 50.04 & 47.07 && 31.63 & 49.57 & 46.87 & 50.09 & 49.97 && 49.01 & 31.63 & 40.26\\
& w & 37.84 & 49.95 & 50.00 & 33.20 & 49.78 & 49.94 & 45.21 & 4.14 && 37.97 & 49.95 & 46.85 && 31.63 & 49.49 & 46.56 & 49.96 & 49.90 && 48.87 & 31.63 & 40.18 \\
\midrule
\multirow{2}{*}{HME~\cite{HME}} & w/o & 36.44 & 49.98 & 50.09 & 32.53 & 49.79 & 50.02 & 42.67 & \textcolor{blue}{6.53} && 36.58 & 50.05 & 45.84 && 29.52 & 49.16 & 46.12 & 50.17 & 49.93 && 48.68 &29.52 & 39.03\\
& w & 37.42 & 49.90 & 49.97 & 33.21 & 49.77 & 49.96 & \textcolor{blue}{47.03} & 5.43 && 37.55 & 49.99 & 47.58 && 31.01 & 49.71 & 46.42 & 49.87 & 49.96 && 48.91 & 31.01 & 39.89\\
\midrule
\multirow{2}{*}{HCRN~\cite{hcrn}} & w/o & 37.78 & \textcolor{blue}{50.12} & 49.99 & 33.62 & 49.78 & 50.10 & 43.66 & 5.15 && 37.90 & 50.11 & 46.22 && 31.24 & 49.29 & 47.36 & 50.21 & 50.11 && 49.12 & 31.24 & 40.11 \\
& w & 40.33 & 49.86 & 49.85 & 33.55 & 49.70 & 50.01 & 43.84 & 5.52 && 40.33 & 49.96 & 46.41 && 36.34 & 49.22 & 43.42 & 50.02 & 50.01 && 47.97 & 36.34 & 42.11 \\

\midrule
\multirow{2}{*}{Ours (100K)} & w/o & 37.42 & 49.94 & 50.06 & 32.53 & 49.77 & 49.97 & 46.62 & 5.06 && 37.57 & 49.96 & 47.27 && 30.92 & 49.66 & 46.69 & 50.01 & 49.97 && 48.98 & 30.92 & 39.88\\
& w & \textcolor{blue}{41.93} & 49.26 & \textcolor{blue}{51.52} & \textcolor{blue}{35.24} & \textcolor{blue}{50.11} & \textcolor{blue}{52.24} & 45.62 & 5.61 && \textcolor{blue}{42.17} & \textcolor{blue}{51.14} & 46.36 && \textcolor{blue}{38.69} & 49.82 & 42.37 & \textcolor{blue}{50.84} & \textcolor{blue}{52.59} && 48.77 & \textcolor{blue}{38.69} & \textcolor{blue}{43.69}\\
\cmidrule{3-24}
\multirow{2}{*}{Ours (full)} & w/o & 38.72 & 50.03 & 49.99 & 33.87 & 49.85 & 50.02 & 48.23 & 5.80 && 38.83 & 50.01 & \textcolor{blue}{48.11} && 32.58 & \textcolor{blue}{49.94} & \textbf{47.96} & 50.16 & 49.98 && \textcolor{blue}{49.43} & 32.58 & 40.95\\
&w &\textbf{46.42} & \textbf{60.67} & \textbf{64.63} &\textbf{38.83} & \textbf{62.17} & \textbf{56.06} & \textbf{48.15} & \textbf{10.12} && \textbf{47.61} & \textbf{56.19} & \textbf{53.83} && \textbf{43.42} & \textbf{60.68} & \textcolor{blue}{47.76} & \textbf{52.86} & \textbf{56.63} && \textbf{55.04} & \textbf{43.42} & \textbf{49.20} \\

\bottomrule
\end{tabular}
\end{table*}

\section{Additional results and analyses} \label{quant_results}
Here, we discuss further results and analyses on AGQA and STAR benchmarks.

\subsection{AGQA}
\subsubsection{Performance comparison w.r.t. training data}
To train on AGQA, we split the AGQA training set into 90\%-10\% train-val split. The new training set after this split comprises approximately 1.4M QA pairs. From this training set, we randomly sampled 100K data samples to train our network. We find the SHG-VQA to outperform the baselines even when trained with 100K samples which is $\sim15\times$ less training data than the data used to train the baseline methods (see fig.~\ref{fig:perform-comp}). More specifically, SHG-VQA obtains 43.69\% vs. 42.11\% for HCRN which is the best model for AGQA on overall VQA accuracy. Similarly, we obtain on par or often better performance on the three testing metrics of indirect references, novel compositions and more compositional steps. We provide a detailed breakdown of our results with 100K and 1.4M training samples in comparison with the baselines which were trained on the full training set of 1.6M QA pairs. See table~\ref{tab:agqamain-table-supp},~\ref{tab:novel-comp-supp},~\ref{tab:agqa-more-comp-steps-supp} for detailed results.
\subsubsection{Results for more compositional steps}
AGQA provides a train-test split to test models's generalization to more compositional steps where training split has questions with fewer compositional steps. On this metric, SHG-VQA with 100K training samples achieves comparable results to the SOTA model. When compared to the best performing model for each question type, our full model gains $\uparrow$ 4.15\% absolute points over the best model (HME: 48.09\% vs. ours: 52.24\%) for binary questions, $\uparrow$ 1.2\% improvement over SOTA (HCRN:23.70\% vs. ours:24.90\%) for open-answer questions, achieving overall $\uparrow$ 4.14\% improvement on all questions. Fig.~\ref{lin_reg_graphs} shows correlation between accuracy and compositional steps. A linear regression model is fit to each method's performance w.r.t number of compositional steps. For \textbf{binary questions}, the baseline methods perform significantly lower even with single compositional-step questions, whereas our model unsurprisingly yields the highest accuracy. SHG-VQA is consistently better for all compositional steps on binary question than the baselines. Nonetheless, we observe a negative correlation between accuracy and compositional steps for binary questions. For \textbf{open} questions, a slightly positive correlation between accuracy and compositional steps is noticed for all methods including SHG-VQA. For overall accuracy on this metric, although SHG-VQA is able to bridge the gap between human accuracy and VQA algorithms by providing SOTA results, there is still large room for improvement on this novel task.

\begin{table}[h]
  \caption{Evaluation on AGQA's \textbf{novel compositions.}}
    \label{tab:novel-comp-supp}
 \renewcommand{\arraystretch}{.9}
  \centering \scriptsize \setlength{\tabcolsep}{.6\tabcolsep}
  {\begin{tabular}{c c c c c} 
  \toprule
  
  Method & training data size & Binary & Open & All \\
  \midrule
  PSAC & 1.6M & 46.49 & 19.34 & 34.71\\
  HME & 1.6M & 45.42 & 17.17 & 33.15 \\
  HCRN & 1.6M & 44.88 & 20.12 & 34.13\\
  \midrule
  SHG-VQA & 100K &\textcolor{blue}{46.55} & \textcolor{blue}{22.2} & \textcolor{blue}{36.01}\\
  SHG-VQA & 1.4M & \textbf{49.27} & \textbf{25.92} & \textbf{39.15}\\
 
    \bottomrule
    \end{tabular}}
\end{table}

\begin{table}[h]
 \renewcommand{\arraystretch}{.9}
  \centering \scriptsize \setlength{\tabcolsep}{.6\tabcolsep}
  \caption{Comparison on AGQA's \textbf{more compositional steps} with our model with 100K training samples and full training set.}
  {\begin{tabular}{c c c c c} 
  \toprule
  Method  & training data size & Binary & Open & All \\
    \midrule
    
    PSAC~\cite{PSAC} & 1.6M & 47.65 & 14.81 & 47.19\\
    HME~\cite{HME} & 1.6M & \textcolor{blue}{48.09} & 20.98 & \textcolor{blue}{47.72}\\
    HCRN~\cite{hcrn} & 1.6M & 46.96 & \textcolor{blue}{23.70} & 46.63 \\
    \midrule
    SHG-VQA & 100K & 47.13 & 22.66 & 46.97 \\
    SHG-VQA  & 1.4M & \textbf{52.24} & \textbf{24.90} & \textbf{51.86} \\
    \bottomrule
    \end{tabular}}
    
    \label{tab:agqa-more-comp-steps-supp}
\end{table}

\begin{table}
\centering
\caption{Additional results on AGQA for all question types.}
  \label{tab:table6}
\resizebox{\columnwidth}{!}
  {\begin{tabular}{r c c c c} 
  \toprule
     Question Types & Blind Model (Q-Only) & Deaf Model (V+HG) & SHG-VQA-100K\\
    \midrule
    
    \multirow{7}{2em}{\rotatebox{90}{Reasoning}} 
    object-relationship & 37.42 & 15.16 & \textbf{41.93}\\
    relationship-action & \textbf{49.94} & 0.01 & 49.26\\
    object-action  & 50.06 & 0.06 &\textbf{51.52}\\
    superlative & 32.53 & 14.88 & \textbf{35.24}\\
    sequencing  & 49.77 & 0.04 &\textbf{50.11} \\
    exists  & 49.97 & 17.91 &\textbf{52.24}\\
    duration comparison  & \textbf{46.62} & 7.89 &45.62 \\
    activity recognition  & 5.06 & 0.00 &\textbf{5.61}\\
    \hline
    
    \multirow{3}{7.1em}{\rotatebox{90}{Semantic}} 
    object  & 37.57 & 13.71 & \textbf{42.17} \\
    relationship & 49.96 & 13.92  & \textbf{51.14}\\
    action  & \textbf{47.27} & 2.92 & 46.36\\
    \hline
    
    \multirow{5}{7.1em}{\rotatebox{90}{Structure}} 
    query  & 30.92 & 15.63 & \textbf{38.69} \\
    compare  & 49.66 & 1.08  & \textbf{49.82}\\
    choose & \textbf{46.69} & 9.72 & 42.37  \\
    logic  & 50.01 & 18.02 & \textbf{50.84} \\
    verify  & 49.97 & 18.12  & \textbf{52.59}\\
    \hline
    
    binary & \textbf{48.98} & 10.65 & 48.77  \\
    open  & 30.92 & 15.63  & \textbf{38.69}\\
    all  & 39.88 & 13.16 & \textbf{43.69}\\
  \bottomrule
  \end{tabular}}
  
\end{table}

\subsubsection{Additional results on AGQA for model variations}
AGQA~\cite{grunde2021agqa} report results for each baseline with language-only model to compare with the respective full models. Following this, we train SHG-VQA in three settings on AGQA: blind model (\textbf{w/o vision}), deaf model (\textbf{vision-only}), and full model. We perform this study using our 100K subset. Results are discussed below:
\noindent \textbf{Blind model performance}
We evaluate our question-only model which is a BERT-like 5 layers transformer encoder against our full vision model (table \ref{tab:table6}) to measure how much linguistic bias our model is able to exploit from the dataset. With results comparable to HCRN’s vision and no-vision counterparts, our language model is able to achieve an overall video-question answering accuracy of 39.88\%, only 3.81\% less than our vision model. The vision model outperforms its language-only counterpart throughout a majority of the question types, however the language-only model has slight improvements over the vision model in duration comparison question types and action semantics, where it performs 1\% better. 
Additionally, the language model also performs slightly better in regards to overall accuracy on binary question types. Further examining binary question categories (table \ref{tab:table6}) show that the model again performs roughly 1\% better than its vision counterpart on binary object-relationship and  duration comparison reasoning categories, as well as binary object and action semantic question types. The most noteworthy difference is that this model outperforms the vision model by 4.32\% in the choose structural category. Overall, the full model outperforms this language-only model in most categories.
\begin{table}
  \caption{\textbf{Results for different training protocols.} Results shown for STAR test set. Rows 1,2, and 3 are with SlowR50  and rows 4,5 show results with MViT-B backbone.}
  \label{tab:ablations}
   \renewcommand{\arraystretch}{.9}
  \centering \scriptsize \setlength{\tabcolsep}{.3\tabcolsep}
  \begin{tabular}{lccccc}
    \toprule
 Q. Type & Interaction & Sequence & Prediction & Feasibility & Overall \\
 \midrule
 (1) separate training & \textbf{47.98} & 42.03 & 35.34 & 32.52 & \textbf{39.47}\\
 (2) all w/ filtered data & 37.67 & 36.91 & 37.29 & \textbf{33.04} & 36.23\\
 (3) all-SlowR50 & 42.38 & \textbf{42.49} & \textbf{37.85} & 30.78 & 38.37\\
 \midrule
 (4) all--SlowR50 & 42.38 & 42.49 & 37.85 & 30.78 & 38.37\\
 (5) all--MViTB & \textbf{43.35} & \textbf{44.37} & \textbf{38.55} & \textbf{33.91} & \textbf{40.04}\\


    \bottomrule
  \end{tabular}
\end{table}
\begin{table}[t]
  \caption{SHG-VQA with SlowR50 backbone evaluated on STAR-Humans. Numbers are reported in percentages.}
  \label{tab:star-humans}
  \renewcommand{\arraystretch}{.9}
  \centering \scriptsize \setlength{\tabcolsep}{.3\tabcolsep}
  \begin{tabular}{lccccc}
    \toprule
 
      &  Interaction & Sequence & Prediction & Feasibility & Overall \\
    \midrule
 SHG-VQA & \textbf{52.00}  & \textbf{45.00} & 31.00 & 23.00  & 37.75\\
    \bottomrule
  \end{tabular}
\end{table}
\noindent \textbf{Deaf model performance}
In addition to the language-only model, we also train a deaf (vision-only) model to measure biases that may arise from the visual input alone (table \ref{tab:table6}). This version of our model obtained an overall VQA accuracy of 13.16\% on all question types, performing worse than both our full, and language-only models in every question category. From this, we conclude that the visual bias is much less than the language bias.

\noindent {\bf Ablation on T/M/N?} We chose clip length T=16 following prior works. We report results for varying clip length T on AGQA dataset in Tab.~\ref{tab:agqa_T} with models are trained for 10-15 epochs on 100K QA pairs. Hyperparameters M and N capture the number of actions and relations we want to predict for each frame. Therefore, video length does not effect M/N. 

\begin{table}[h]
  \vspace{-5pt}
  \renewcommand{\arraystretch}{.9}
  \centering \scriptsize \setlength{\tabcolsep}{.25\tabcolsep}
  \begin{tabular}{lccc}
    \toprule
{\bf T} &  {\bf binary} & {\bf open} & {\bf all} \\
\midrule
16 & 48.77 & 38.69 & 43.69 \\
24 & 46.30 & 38.89 & 42.57\\
32 & 45.50 & 37.29 & 41.36 \\
\bottomrule
\end{tabular}
\caption{Results on AGQA dataset for varying video clip length T. 
}
\label{tab:agqa_T}
\end{table}

\subsection{STAR}
\subsubsection{Results w.r.t different training protocols} We experimented with different training protocols for STAR dataset including separate trainings used in ~\cite{wu2021star}, single model with filtered questions as explained in ~\ref{train-protocol}, and training a single model on the full training set. We use SlowR50 backbone for this study and find that using separate trainings is most beneficial for interaction questions and overall accuracy. Using filtered questions although perform best for feasibility questions, but it hurts the performance on other question types. When trained on full training data with SlowR50 and MViT-base backbones, using MViT yields better performance. 

\begin{table}[h]
   \renewcommand{\arraystretch}{.9}
  \centering \scriptsize \setlength{\tabcolsep}{.3\tabcolsep}
  \begin{tabular}{lccccc}
    \toprule
    
    Set Pred. Loss &  Interaction & Sequence & Prediction & Feasibility & Overall \\
    \midrule
full video & 39.81 &	40.69&	30.17&	29.91&	35.15 \\
frame-wise &39.42	& 41.83	& 33.8 &	27.48 &	35.63 \\
   \bottomrule
  \end{tabular}
  \vspace{-10pt}
   \caption{Results for SHG-VQA model on STAR test set. }
  \label{tab:quant}
\end{table}

\subsection{Results on STAR-Humans}
STAR-Humans is a subset provided by ~\cite{wu2021star} with 400 free-form questions asked by humans. We evaluate SHG-VQA-SlowR50 on STAR-Humans and obtain the results shown in table~\ref{tab:star-humans}. For this subset, SHG-VQA performs best for interaction questions (52.00\%) and worst on feasibility questions (23\%). 

\begin{table*}[t]
  \caption{Training configurations for SHG-VQA on STAR dataset.}
  \label{tab:training-details}
   \renewcommand{\arraystretch}{.9}
  \centering \scriptsize \setlength{\tabcolsep}{.3\tabcolsep}
  \begin{tabular}{lcccccc}
    \toprule
 
    Experiment   & backbone &  Batch Size & Models trained & Loss & Test set & Overall Acc.\\
    \midrule
    \underline{\textit{Adapted batch size per QT:}} \vspace{5pt} \\
    
 SHG-VQA (Q + HG) & Slow\_R50 & (16, 16, 16, 16) & 4 &  $L (eq. 1)$ & test & 39.47\\
 SHG-VQA (Q + HG) & Resnext101 & (16, 16, 4, 4) & 4 &  $L (eq. 1)$ & test & 38.28\\
SHG-VQA (Q + V) & Slow\_R50 & (8, 8, 8, 8) & 4 & $L_{vqa}$ & test & 30.81\\
SHG-VQA (Q + HG) & Slow\_R50 & (16, 16, 16, 16) & 4 &  $L (eq. 1)$ & test & 39.47\\
SHG-VQA (Q + V + HG) & Slow\_R50 & (16, 16, 16, 16) & 4 & $L (eq. 1)$ &test & 39.18\\
\midrule
    \underline{\textit{Hypergraph components:}} \vspace{5pt} \\
    
SHG-VQA (Q + HG) Action only -- Act=3  &  Slow\_R50 & 32 & 1 & $L (eq. 1)$ &val & 38.68\\
 SHG-VQA (Q + HG) Rel. only -- Rel=8 &  Slow\_R50 & 32 & 1 & $L (eq. 1)$ & val &35.16\\
 SHG-VQA (Q + HG)   Both -- Act=3, Rel=8 & Slow\_R50 & 32 & 1 & $L (eq. 1)$ & val &39.20\\
    \midrule
    \underline{\textit{Number of queries}} \vspace{5pt} \\
    
  SHG-VQA (Q + HG)  Act=2, Rel=8 & Slow\_R50 & 32 & 1 & $L (eq. 1)$ &val & 38.34\\
  SHG-VQA (Q + HG)  Act=3, Rel=8 & Slow\_R50 & 32 & 1 & $L (eq. 1)$ &val & 39.20\\
  SHG-VQA (Q + HG)  Act=4, Rel=8 & Slow\_R50 & 32 & 1 & $L (eq. 1)$ &val & 37.06\\
  SHG-VQA (Q + HG)  Act=3, Rel=12 & Slow\_R50 & 32 & 1 & $L (eq. 1)$ &val & 39.90\\
  SHG-VQA (Q + HG)  Act=4, Rel=12 & Slow\_R50 & 32 & 1 & $L (eq. 1)$ &val & 38.39\\

    \bottomrule
  \end{tabular}
\end{table*}

\begin{figure*}[t]
\begin{center}

\includegraphics[width=\linewidth]{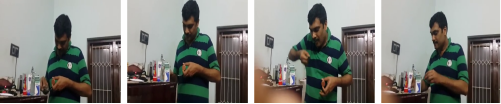}\\
Sample video\\
  \includegraphics[width=\linewidth]{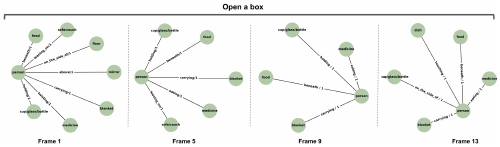}\\
  Groundtruth hypergraph\\
  \vspace{15pt}
  \includegraphics[width=\linewidth]{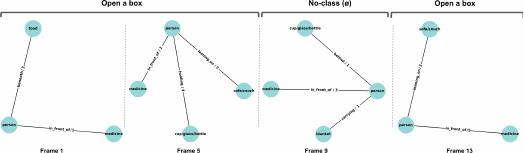}\\
  Optimal matching over full video\\
  \vspace{15pt}
  \includegraphics[width=\linewidth]{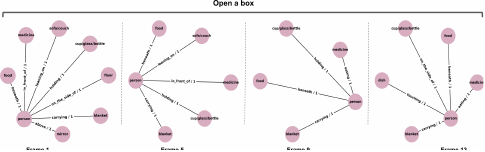}\\
  Optimal matching for each timestep $t$ (proposed in eq.2 and eq.4 in the main paper)\\
 \end{center}
  \caption{Ground-truth and predicted situation hyper-graph for every 4th frame in a clip of length 16. Row 1 shows video frames, row 2 shows the ground-truth situation hyper-graph, row 3 shows predicted graphs from the model with set prediction loss without considering frames, row 4 shows predicted hyper-graph when the model is trained by matching each timestep $t$ (the proposed loss function). The edges show the person-object relationship labels along with the number of times it was predicted. (see Section~\ref{sec:qual_results-supp} for discussion about results.)} \label{fig:qual_gt-1}
\end{figure*}

\begin{figure*}[t]
\begin{center}
\includegraphics[width=\linewidth, height=1.5in]{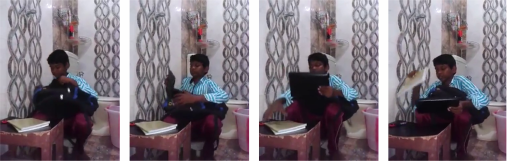}\\
Sample video\\
  \includegraphics[width=\linewidth]{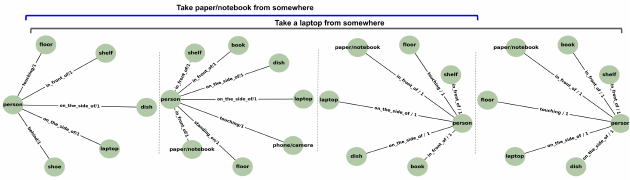}\\
  Groundtruth hypergraph\\
  \vspace{15pt}
  \includegraphics[width=\linewidth]{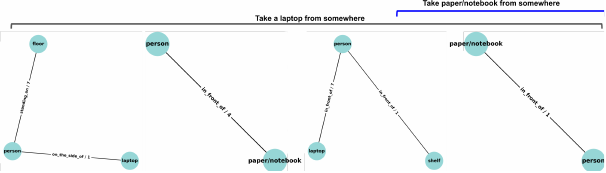}\\
  Optimal matching over full video\\
  \vspace{15pt}
  \includegraphics[width=\linewidth]{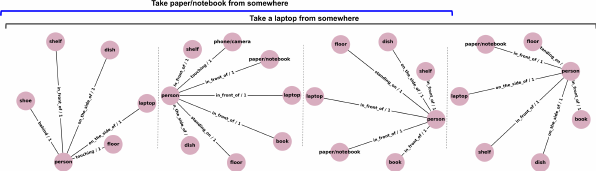}\\
Optimal matching for each timestep $t$ (proposed in eq.2 and eq.4 in the main paper)\\
 \end{center}
  \caption{\small Ground-truth and predicted situation hyper-graph for every 4th frame in a clip of length 16. Row 1 shows video frames, row 2 shows the ground-truth situation hyper-graph, row 3 shows predicted graphs from the model with set prediction loss without considering frames, row 4 shows predicted hyper-graph when the model is trained by matching each timestep $t$ (the proposed loss function). The edges show the person-object relationship labels along with the number of times it was predicted.  (see Section~\ref{sec:qual_results-supp} for discussion about results.)} \label{fig:qual_gt-2}
\end{figure*}

\section{Qualitative results}
\label{sec:qual_results-supp}

\noindent \textbf{Comparison between optimal matching with full video compared to optimal matching for each timestep:} Figures~\ref{fig:qual_gt-1} and \ref{fig:qual_gt-2} show qualitative comparison of predicted situation hyper-graphs from situation hyper-graph decoder in the proposed model. Note that the situation hyper-graph solely relies on the video input. Hence, we show the input video, ground-truth graph, and predicted situation hyper-graphs in two settings: 
1) optimal matching with full video instead for each timestep $t$ over $L_{Act}$ and $L_{rel}$ (baseline); 
2) optimal matching for each timestep $t$ for the actions and relations set predictions (as described in $eq. 2$ and $eq. 4$ in the main paper). 
%
We observe that when using the optimal matching without imposing the time constraint (i.e., to do optimal matching at each time step), it results in duplicate predictions at frame level. In figure~\ref{fig:qual_gt-1} and ~\ref{fig:qual_gt-2}, the first two rows show the video frames and the corresponding ground truth situation hyper-graph respectively. Row 3 shows the situation hyper-graphs we obtain with the optimal matching for the full video. In edge labels, we show the predicted relationship as well as the count of multiple edges between two nodes. For brevity, we show every $4^{th}$ frame i.e., frames 1, 5, 9, 13. 
We can see in row 3, that the predicted graph is sparse and not able to capture all relationships due to suffering from the duplicate predictions problem. Additionally, it sometimes predicts no-class $\phi$ label i.e. empty set for actions and predictions as we can see in fig~\ref{fig:qual_gt-1}, row 3, column 3. 
Row 4 shows the predicted situation hyper-graphs with the imposed constraint of optimal matching at frame-level. We can see that the proposed solution for optimal matching greatly improves the quality of generated hyper-graphs. See Tab.~\ref{tab:quant} for quantitative results.

\section{Computational Cost of SHG-VQA}
\label{sec:computational-cost}
The computational cost of SHG-VQA includes video and text encoders with the little overhead from decoders for decoding graph queries.  
At inference time, the decoders' output is directly sent to cross-attentional transformer along with the meta embeddings without any graph prediction. Given that we only use L=5 layers for all encoders and decoders in SHG-VQA, the depth of the SHG-VQA is 12 layers i.e., comparable to existing vision-language methods~\cite{ALBEF,khan2022_wsg_vlt}, e.g., ALBEF.

\section{Ethical considerations} \label{sec:limitations}

As our system is trained on real-world data, it might capture negative data inherent biases, such as actions only executed by people with specific clothing or stereotype questions. We are not aware of such stereotypes in the here used datasets AGQA and STAR but would recommend assessing the fairness of any system based on this work before putting it in any production environment.



\end{document}